\documentclass[11pt]{article}

\PassOptionsToPackage{table}{xcolor}
\usepackage[preprint]{acl}

\usepackage{times}
\usepackage{latexsym}
\usepackage{amsmath}
\usepackage{amssymb}

\usepackage{float}

   \usepackage{graphicx,array}  
   \usepackage{booktabs}
\definecolor{proposedbg}{HTML}{E8F0FB}

\usepackage[T1]{fontenc}

\usepackage[utf8]{inputenc}

\usepackage{microtype}

\usepackage{inconsolata}

\usepackage{graphicx}

\usepackage{booktabs}
\usepackage{tabularx}

\usepackage{xspace}

\NewDocumentCommand{\yunfan}{s m}{%
  \IfBooleanF{#1}{{\bfseries\color{blue} Yunfan: #2}}}
\NewDocumentCommand{\smara}{s m}{%
  \IfBooleanF{#1}{{\bfseries\color{violet} Smara: #2}}}
\NewDocumentCommand{\kmnote}{s m}{%
  \IfBooleanF{#1}{{\bfseries\color{red} Kathy: #2}}}

\usepackage{appendix_boxes}
\definecolor{appendixaccent}{RGB}{65,84,102}
\newenvironment{appendixtext}{%
  \par\addvspace{6pt}%
  \normalfont\normalsize\raggedright
  \setlength{\parindent}{0pt}\setlength{\parskip}{0pt}%
  \setlength{\leftskip}{6pt}\setlength{\rightskip}{6pt plus 1fil}%
  \appendixboxstart
}{\par\appendixboxend\addvspace{14pt}}
\newcommand{\excerptlabel}[1]{%
  \par\addvspace{3pt}{\color{appendixaccent}\bfseries
    \interlinepenalty=10000 [#1]\par}\nobreak}
\newcommand{\promptlabel}[1]{%
  \par\addvspace{3pt}[#1]\par\nobreak}

\newcommand{\projectname}[0]{\textsc{Memento}\xspace}

\title{\projectname: Learning Context Management for Long Context Processing Agents}

\author{
  \textbf{Yunfan Zhang},
  \textbf{Taehoon Hwang},
  \textbf{Kathleen McKeown},
  \textbf{Smaranda Muresan}
  \\
  \\
  Columbia University
  \\
  New York, NY, USA
  \\
  \texttt{yunfan.z@columbia.edu}
}

\begin{document}
\maketitle
\begin{abstract}
LLM agents that operate over long inputs must perform \emph{context management}: deciding what to retain in a fixed-size context window. A common approach is compaction, where the model replaces its history with a self-generated note. Yet off-the-shelf agentic models, including GPT-5.5 and GLM-5.3, often write ineffective notes, and reinforcement learning (RL) alone rarely discovers useful note-taking behavior. We present \projectname{}, a long-context agent built on Qwen3-8B and trained to manage its context through compaction. We first bootstrap note-taking by fine-tuning on synthetic trajectories annotated with notes generated in hindsight, then reinforce it with RL using a reward that combines answer correctness, note-quality heuristics, and confidence calibration. Although trained only to compact into a fresh window, \projectname{} extends at test time to strategies such as context folding and confidence-based backtracking. On book-level question answering across four benchmarks, with books of up to 2.5M tokens, \projectname{} raises Qwen3-8B's accuracy from 36.9\% to 67.8\% while never holding more than 20K tokens in context. It outperforms GPT-5.5 agent in the same harness (13.1\%) and full book context (56.7\%) at less than 1/50 the cost. It also nearly matches a $\sim$100$\times$ larger GLM-5.3 agent (69.3\%) at 1/9 the cost, and comes within 6.4 points of GLM-5.3 full context at 1/26 the cost.
\end{abstract}

\section{Introduction}
\label{sec:introduction}
\begin{figure}[h]
    \centering
    \includegraphics[width=0.95\columnwidth]{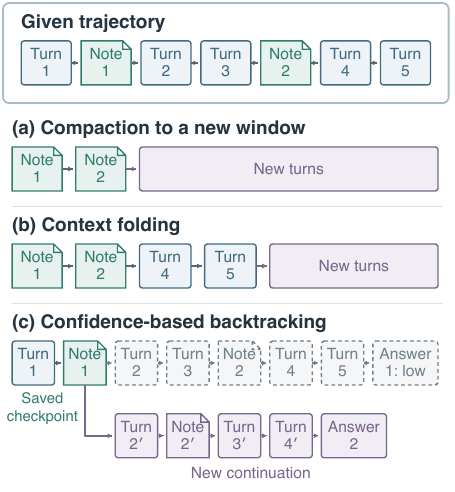}
    \caption{Context management methods incorporated in this work. (a) Compaction starts a new window containing only the saved notes and discards all prior turns. (b) Context folding additionally retains the turns after the latest note. (c) Confidence-based backtracking resumes from an earlier note checkpoint after a low-confidence answer and generates an alternative continuation.}
    \label{fig:compaction_methods}
\end{figure}

Large language model (LLM) agents are taking on ever longer tasks, and the horizon over which they can work autonomously has grown steadily~\citep{kwa2025measuring}.
This makes agents a natural fit for long-context processing: rather than ingesting an entire book or knowledge base at once, an agent can search the input and read only the parts it needs.
Most agents, however, follow the ReAct paradigm~\citep{yao2023react}, appending every thought, action, and observation to their context.
Because the history grows with the task while the context window stays fixed, a sufficiently long task will eventually exhaust the window.
Larger context windows only postpone the problem: LLMs underuse information in long inputs~\citep{liu2024lost}, and their performance can degrade as the context grows~\citep{du2025context}.
Deciding what an agent should retain, which we call \textit{context management}, is therefore central to agentic long-context processing.

Prior approaches either prompt the model to manage its own memory or delegate summarization to a separate component \citep{packer2023memgpt,wu2025resum}.
Another line of work learns context management through reinforcement learning or supervised fine-tuning \citep{zhou2025mem1,yu2025memagent,sun2025folding,ye2025agentfold}.
\smara*{I feel the above is too detailed for intro. so maybe "Prior approaches either prompt the model to manage its own memory or delegate summarization to a separate component \cite{} (put all citations there). Another line of work learns context management through post-training either via RL or supervised-fine tuning \cite{}." And leave the individual work for related work?}

In this work, we jointly design the training pipeline and the test-time harness for context management in agentic long-context processing.
We take book-level multi-hop question answering (Book QA) as our target task, since answering a single question often requires gathering evidence scattered across an entire book.
Alongside standard search and view tools, the agent has a \textit{compaction note} tool that it may call at any point to summarize its progress, record mistakes, and plan next steps.
Notes persist through every context management operation we consider (Figure~\ref{fig:compaction_methods}) and double as checkpoints for backtracking.
\kmnote*{Well now that I've said I like the figure, in looking back I'm a little confused: how do notes persist? My understanding is that when you summarize they disappear? Or is that each note is a summary and replaces the chain? Actually I think the latter. NO, looking at the figure again, I understand. Might help to say somewhere that a note is a summary that replaces the turns it summarizes. }

We present \textbf{\projectname}, a long-context processing agent built on Qwen3-8B~\citep{qwen3technicalreport}.
Off-the-shelf LLMs, including stronger models such as GPT-5.5 and GLM-5.3, never make valid calls to our compaction note tool to record progress or plan ahead, even when explicitly introduced in their system prompt, leaving RL with little useful note-taking behavior to reinforce.
\kmnote*{Is this calls to your tool or their own? Unclear. And if yours, how do they know about it? } 
We therefore shape the initial policy with synthetic SFT data.
We inject compaction notes post hoc into correct GLM-5.3~\citep{zai2026glm53} agent rollouts and fine-tune on the resulting synthetic trajectories.
We further train the SFT-initialized policy with RL, using a reward that combines format compliance, heuristics on note quality, final-answer correctness, and, optionally, the calibration of a verbalized confidence score.

Throughout training, the agent uses only the simplest operation, compaction into a new window (Figure~\ref{fig:compaction_methods}a), carrying its notes across multiple windows within a single rollout.
Because context folding (b) and confidence-based backtracking (c) operate on the same notes and prompts, we can adopt them at test time with minimal train--test mismatch, and choose among these methods according to the accuracy and budget of a given deployment.

We train and evaluate \projectname{} on LiteraryQA~\citep{bonomo2025literaryqa}, NovelHopQA~\citep{gupta2025novelhopqa}, and the English subset of SagaScale~\citep{du2026sagascale}, and additionally evaluate on NovelQA~\citep{wang2025novelqa}, which is held out from training.
On 4.7K test questions, with accuracy micro-averaged over the four datasets, our full system reaches 67.8\%.
This greatly improves the 36.9\% of the original Qwen3-8B agent and nearly matches the 69.3\% of a $\sim$100$\times$ larger GLM-5.3 agent, with all agents run in the same harness using context folding and confidence-based backtracking.
The conventional long-context approach, placing the entire book in GLM-5.3's context window, yields 74.3\%, 6.4 points higher, but at more than 26$\times$ the inference cost (in USD).
\projectname{} also outperforms GPT-5.5, which reaches only 13.1\% in our harness and 56.7\% even with the full book in its context window, at 54$\times$ our cost.
These results suggest that a small open model agent can process long-context inputs effectively when context management is learned jointly with the task and the harness.

Our contributions are as follows:
\begin{itemize}
    \item \textbf{SFT and RL for learned context management.} We bootstrap note-taking with SFT on agentic rollouts augmented with synthetic compaction notes, then refine it with RL under a hybrid reward that incorporates process signals on note quality with outcome rewards on answer correctness. Together, they improve the Qwen3-8B agent from 36.9\% to 66.3\%.
    \item \textbf{Confidence-guided backtracking.} We reward calibrated verbalized confidence during RL and use it at test time to backtrack over saved notes and retry low-confidence answers, raising accuracy further to 67.8\%.
    \item \textbf{Co-designed training and test-time harness.} The agent can call the compaction note tool at any point in its trajectory, so the same note structure allows several test-time strategies: compaction into a new window, context folding, and backtracking to earlier notes. Although the agent is trained only with compaction into a new window, it transfers to folding and backtracking with minimal train--test mismatch, so the harness can be chosen to fit a deployment's cost--accuracy trade-off.
\end{itemize}

\section{Related Works}
\label{sec:related_works}

\paragraph{Long-document question answering and reading agents.}
Question answering over entire books has long served as a test of deep narrative comprehension.
To tackle these problems, prior work such as MemWalker~\citep{chen2023memwalker} organizes a text into a tree of summaries that the model navigates through iterative prompting, ReadAgent~\citep{lee2024readagent} compresses episodes into gist memories and looks up original passages when needed, and GraphReader~\citep{li2024graphreader} lets an agent explore a graph built from the document.
These systems use document summaries or graphs built before question answering, whereas our agent searches the raw text and compacts its context during question answering, using a policy learned through training. \kmnote*{Could you also contrast by saying "; it also uses inference time compaction {\em during} question answering." (Because I assume training is before QA right"?}

\paragraph{Context management and compaction for agents.}
Context management approaches that can be applied without updating the agent's policy parameters include hierarchical memory paging~\citep{packer2023memgpt} and ReSum's inference-time use of a separate summarization tool~\citep{wu2025resum}.
Complementary work on long-term conversational memory includes MemGuard~\citep{ha2026memguard}, which separates memories by functional type and selectively retrieves them to reduce memory contamination.
Other approaches train compaction with RL, either as a compact internal state updated at every turn~\citep{zhou2025mem1}, a fixed-length memory overwritten while reading a document segment by segment~\citep{yu2025memagent}, or summaries and folded sub-trajectories optimized end to end with the policy~\citep{lu2025supo,sun2025folding}.
AgentFold~\citep{ye2025agentfold} learns multi-scale folding operations through supervised fine-tuning alone.
FoldAct~\citep{shao2025foldact} also initializes with SFT on summary-augmented trajectories before RL, focusing on joint optimization of summaries and actions.
Our pipeline inserts compaction notes into teacher trajectories in hindsight, then refines the initialized policy with RL.
\kmnote*{Could you refer to this paper where you think it is appropriate. This is work with Heng Ji's group and I'm concerned that one of them could be a reviewer: https://arxiv.org/abs/2605.28009}


\paragraph{Confidence and test-time search.}
Spending more computation at inference can improve accuracy, through majority voting over sampled reasoning paths~\citep{wang2023selfconsistency}, tree search with lookahead and backtracking~\citep{yao2023tot}, or retries guided by verbal self-reflection~\citep{shinn2023reflexion}.
However, deciding when to spend this computation requires a signal of reliability.
Verbalized confidence from RLHF-tuned models is often better calibrated than their token probabilities~\citep{tian2023justask}, and RLCR~\citep{damani2025rlcr} augments a binary correctness reward with a negative Brier-score term, improving calibration while maintaining comparable accuracy.
We follow the same Brier-style form, and use confidence to decide when a long-horizon agent should backtrack to an earlier note checkpoint and investigate again.

\section{Methodology}
\label{sec:methodology}

\begin{figure*}[t]
    \centering
    \includegraphics[width=\textwidth]{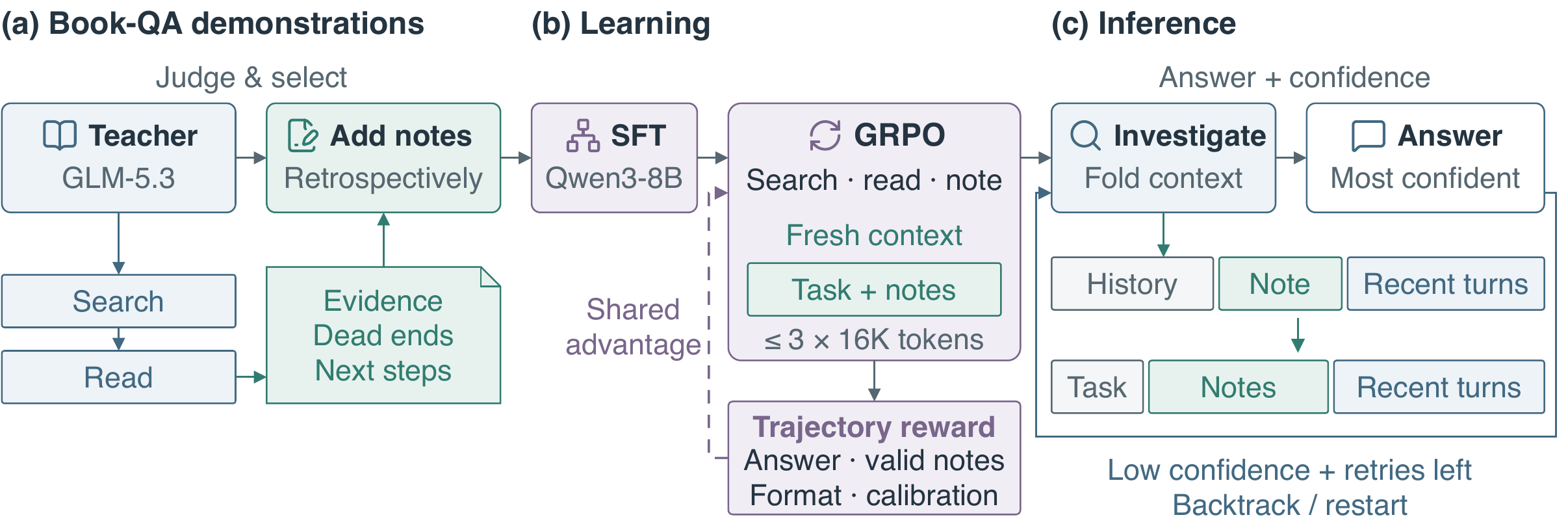}
    \caption{Method overview. (a) We inject note-taking actions into collected demonstrations. (b) SFT initializes the policy, and GRPO reinforces its actions across context resets with a shared trajectory advantage. (c) At inference, context folding retains notes and recent turns, and confidence guides retries and answer selection.}
    \label{fig:methodology}
\end{figure*}

We jointly design the training pipeline and the test-time harness of \projectname{} around the compaction note tool (Figure~\ref{fig:methodology}).
We first shape the initial policy with SFT on agent traces augmented with synthetic compaction notes, which demonstrates effective note-taking strategies to the model.
We then reinforce these strategies with RL using a hybrid reward that involves format compliance, note validity, final-answer correctness, and, optionally, the calibration of a verbalized confidence score, under the compact to new context strategy.
At test time, the same notes support compaction, context folding, and confidence-based backtracking.

\subsection{Task and datasets}
\label{sec:task_dataset}
Given a book and a question about it, the agent must answer in a few words, reading the book only through search tools within a small context window.
Books in our datasets are long: 90\% exceed 51K Qwen3 tokens, 70\% exceed 90K, and the longest reaches 2.5M.
This is far beyond the 16--20K context windows of the models we train, so a Book QA agent must carry its progress across many context windows.
Unlike traditional multi-hop QA, where the supporting evidence amounts to a few paragraphs or articles~\citep{yang2018hotpotqa,ho2020constructing,trivedi2022musique}, Book QA requires evidence scattered across an entire book.
Unlike agentic web QA~\citep{mialon2024gaia,wei2025browsecomp}, it also requires no paid search APIs, whose cost would make RL training infeasible within our budget.

We train and evaluate on LiteraryQA~\citep{bonomo2025literaryqa}, NovelHopQA~\citep{gupta2025novelhopqa}, and the English subset of SagaScale~\citep{du2026sagascale}, and additionally evaluate on NovelQA~\citep{wang2025novelqa}, which is held out from training.
Over 30\% of the books in NovelQA are not publicly released, so they are unlikely to have been memorized during pretraining.
We split each training dataset 8:1:1 into training, validation, and test sets, with no book or question shared across splits.
This yields 17K training questions over 600 books, 2.3K validation questions over 70 books, and 4.7K test questions over 160 books.
All answers are free-form and are graded by an LLM judge against a ground truth of a few words or a short sentence, using SimpleQA prompt (reproduced in \ref{sec:correctness_prompt}).

\subsection{Tools for long-context processing}
\label{sec:tools_long_context}

The book is never prefilled in the context window; the agent accesses it only through three tools.
\texttt{semantic\_search} queries the book in natural language using \texttt{bge-small-en-v1.5} embeddings~\citep{xiao2024cpack} over sliding-window sections of 128--256 tokens, and \texttt{ripgrep\_search} matches exact keywords and regular expressions.
Both return a requested number or page of results with line references, and the agent may then call \texttt{view} to read the full text of any line range.

\subsection{Compaction note tool}
\label{sec:compaction_notes}

Central to our approach, the agent can call the compaction note tool (\texttt{create\_notes}) at any point to record its findings, mistakes, and next steps.
Writing a note does not reset the context, so the agent decides what to preserve as the trajectory unfolds and may write several notes along the way.
The tool keeps a set number of recent notes, each subject to a length limit (Table~\ref{tab:generation_hyperparameters}).
\kmnote*{How is configurable? YOu decide ahead of time or is it automatically configured? }

We consider the notes in a rollout valid only if they satisfy three heuristics:
\begin{itemize}
    \item \textbf{Frequency.} Every nonfinal context window contains at least one note, so that progress is recorded before each reset, and the rollout contains fewer than five notes in total, to discourage excessive note-taking.
    \item \textbf{Placement.} No note is the first tool call of its window or immediately follows another \texttt{create\_notes} call, so that each note covers new work.
    \item \textbf{Diversity.} No two notes have a token set similarity above 80\%, so that the notes cover different parts of the trajectory.
\end{itemize}
During RL training, the note validity reward is assigned only if every note in the trajectory is valid. At test time, after each context management operation, the agent is shown all saved notes, subject to retention and length limits, to maximize the reuse of recorded information.

\subsection{SFT policy initialization}
\label{sec:teacher_sampling}

Neither Qwen3-8B nor stronger models (GPT-5.5, GLM-5.3) reliably write valid notes in our harness, even when the note tool is described in the system prompt.
GPT-5.5 never wrote a compaction note on the 337-sample test set, and none of 12K pilot GLM-5.3 rollouts on the training set contained a valid note. Qwen3-8B, in contrast, tends to write identical notes back to back without intervening tool calls, so only about 6\% of its trajectories contain valid notes.

We therefore initialize the policy with synthetic SFT data from GLM-5.3~\citep{zai2026glm53}, a leading open agentic model that exposes its raw chain of thought~\citep{aa2026glm53}.
We run GLM-5.3 in our harness with compaction into a new context window, as in training, and retain correct trajectories, prioritizing those spanning multiple windows, plus a small number of failed attempts that appropriately expressed low confidence.
We then strip any original notes and prompt GLM-5.3 to insert notes post hoc after major tool-use turns or discoveries, instructing it to summarize key findings, mistakes to avoid, and next steps.
Notes failing the heuristics in Section~\ref{sec:compaction_notes} are resampled; search actions and final answers are left unchanged.

Finally, we fine-tune Qwen3-8B on the resulting ${\sim}$6K trajectories.
As Section~\ref{sec:abl_sft_rl} shows, SFT is not meant to improve accuracy on its own.
Instead, it instills a note-taking strategy that teaches the model when to take notes; RL then reinforces this strategy, yielding higher accuracy.

\subsection{Reinforcement learning}
\label{sec:post_training}

\paragraph{Reward Models.} We use Qwen3-14B as the judge during training and Qwen3-235B-A22B-Instruct-2507 for evaluation.
We intentionally evaluate with a different judge so that reported accuracy is not inflated by any exploitation of the training judge.
In early experiments, we observed frequent reward hacking against the LLM judge during RL training.
Therefore, to detect reward hacking, we first screen each answer for hedging, ambiguity, special tokens, and excessive length, using heuristics and a separate LLM judge call.
Answers that pass the screen are graded by a correctness judge using the SimpleQA prompt~\citep{wei2024simpleqa} with CoT (Appendix~\ref{sec:appendix_prompts}).

\paragraph{Trajectory-level reward.}
We combine four rewards: format, note validity, correctness, and calibration.
The format indicator $F\in\{0,1\}$ is 1 if the final answer and verbalized
confidence are parsable and no reward hacking is detected. It earns a bonus
and gates all other rewards.
$M\in\{0,1\}$ marks valid note-taking, $Y\in\{0,1\}$ marks a correct
final answer, and $p\in[0,1]$ is the value of the verbalized confidence tier.
Following \citet{damani2025rlcr}, we reward calibration with a negated
Brier score, giving the total reward

{\small
\begin{equation}
  R = \lambda_F F + F \bigl( Y + \lambda_M M + \lambda_C [1-(Y-p)^2] \bigr)
\end{equation}
}

\paragraph{Advantage calculation across context windows.}
We use GRPO~\citep{shao2024deepseekmath} to compare trajectories sampled for the same question.
For a question $q$, we sample a group of $G$ trajectories $\{\tau_i\}_{i=1}^{G}$ with rewards $\{R_i\}_{i=1}^{G}$ and assign each trajectory a single advantage
\begin{equation}
    \scalebox{0.7}{$\displaystyle
    \hat{A}_i=\frac{R_i-\operatorname{mean}\bigl(\{R_j\}_{j=1}^{G}\bigr)}{\operatorname{std}\bigl(\{R_j\}_{j=1}^{G}\bigr)}.
    $}
    \label{eq:advantage}
\end{equation}
Let $\mathcal{O}_i$ denote the assistant tokens of $\tau_i$ across all of its context windows, where token $o_{i,t}$ is generated from its window context $c_{i,t}$, and define the importance ratio
\begin{equation}
    \scalebox{0.7}{$\displaystyle
    r_{i,t}(\theta)=\frac{\pi_\theta(o_{i,t}\mid c_{i,t})}{\pi_{\theta_{\mathrm{old}}}(o_{i,t}\mid c_{i,t})}.
    $}
    \label{eq:ratio}
\end{equation}
We maximize
\begin{equation}
    \scalebox{0.7}{$
    \begin{aligned}
        \mathcal{J}(\theta) &= \mathbb{E}_{q,\,\{\tau_i\}}\Biggl[\frac{1}{\sum_{i=1}^{G}|\mathcal{O}_i|}\\
        &\quad \sum_{i=1}^{G}\sum_{t\in\mathcal{O}_i}\Bigl\{\min\bigl(r_{i,t}(\theta)\hat{A}_i,\\
        &\qquad \operatorname{clip}\bigl(r_{i,t}(\theta),1-\epsilon,1+\epsilon\bigr)\hat{A}_i\bigr)\\
        &\qquad -\beta\,\mathbb{D}_{\mathrm{KL}}\bigl[\pi_\theta\,\|\,\pi_{\mathrm{SFT}}\bigr]\Bigr\}\Biggr],
    \end{aligned}
    $}
    \label{eq:grpo_objective}
\end{equation}
where the expectation is over questions $q\sim\mathcal{D}$ and groups $\{\tau_i\}_{i=1}^{G}\sim\pi_{\theta_{\mathrm{old}}}(\cdot\mid q)$, $\epsilon$ is the clipping range, $\mathbb{D}_{\mathrm{KL}}$ is the per-token KL divergence from $\pi_\theta$ to the SFT initialization $\pi_{\mathrm{SFT}}$ at context $c_{i,t}$, and $\beta$ weights this penalty.
Because every assistant token inherits its trajectory's advantage, note-taking actions in earlier windows receive a learning signal from the trajectory's final outcome.

\subsection{Test-time context management}
\label{sec:confidence_search}
As illustrated in Figure \ref{fig:compaction_methods}, all three context management methods, compacting into a new window, context folding, confidence-based backtracking share the same note-taking, prompts, and context layout, in which the task and saved notes always form the prefix of the context window; they differ only in what follows this prefix.
A policy trained only with compaction therefore sees a familiar context under each method, which keeps the train--test mismatch minimal.

\paragraph{Compaction into a new window.}
When the agent reaches the context budget or requests a restart, the conversation history is replaced with a new context window containing only the original task, the tool instructions, and the saved notes. Anything the agent has not yet recorded is lost. This is also the only mode we used to perform RL training of agents.

\paragraph{Context folding.}
Context folding replaces the conversation up to and including the latest note-taking turn with the task and saved notes, while preserving subsequent tool calls, tool outputs, and CoTs verbatim.
The agent thus retains recent work it has not yet summarized, which improves accuracy and token efficiency.

\paragraph{Confidence-based backtracking.}
Each note-taking turn marks a state from which the agent can resume.
When an answer's confidence is at or below a threshold, the agent discards its later work and generates an alternative continuation: if the attempt contains $n$ notes, the first retry attempt resumes right after note $\lfloor n/2\rfloor$. The second retry starts right after the first note. Further attempts restart from scratch.
Backtracking stops once an answer exceeds the confidence threshold; otherwise, we return the highest-confidence answer.
Table~\ref{tab:generation_hyperparameters} gives the thresholds and retry schedule. A sample agent trajectory is provided in Appendix~\ref{sec:trajectory_examples}. 

\begin{table*}[t]
    \centering
    \small
    \setlength{\tabcolsep}{4pt}
    \renewcommand{\arraystretch}{1.08}
    \resizebox{\textwidth}{!}{%
    \begin{tabular}{@{}l<{\hspace{6pt}} l<{\hspace{6pt}} l<{\hspace{6pt}} ccccc rr@{}}
        \toprule
        & & & \multicolumn{5}{c}{Accuracy (\%)} & \multicolumn{2}{c}{Efficiency} \\
        \cmidrule(lr){4-8} \cmidrule(l){9-10}
        Model & Category & Harness & LitQA & NHopQA & NovelQA & Saga & Micro Avg & Tokens & Cost (\$) \\
        & & & {\footnotesize ($n$\,=\,1{,}940)} & {\footnotesize (426)} & {\footnotesize (2{,}305)} & {\footnotesize (37)} & {\footnotesize (4{,}708)} & & \\
        \midrule
        \multicolumn{10}{@{}l}{\textbf{Off-the-shelf models, full book in context}} \\[3pt]
        Qwen3-8B (40K ctx)          & Reference & Full Context & 44.8 & 20.4 & 29.8 & 2.7  & 34.9 & 33K  & 11 \\
        GPT-5.5 High (232K ctx)$^\dagger$ & Reference & Full Context & 51.0 & 54.0 & 67.0 & 51.4 & 56.7 & 144K & 2{,}577 \\
        GLM-5.3 High (1M ctx)       & Reference & Full Context & 76.6 & 75.8 & 71.9 & 75.7 & 74.3 & 167K & 1{,}270 \\
        \midrule
        \multicolumn{10}{@{}l}{\textbf{Off-the-shelf models, context folding agent with confidence-based backtracking}} \\[3pt]
        Qwen3-8B          & Ablation  & Fold + Backtrack & 35.2 & 29.1 & 39.8 & 37.8 & 36.9 & 15K & 37 \\
        GPT-5.5 High$^\dagger$ & Reference & Fold + Backtrack & 10.0 & 9.0  & 19.2 & 16.2 & 13.1 & 84K & 2{,}606 \\
        GLM-5.3 High       & Reference & Fold + Backtrack & 71.2 & 43.4 & 72.1 & 91.9 & 69.3 & 47K & 448 \\
        \midrule
        \multicolumn{10}{@{}l}{\textbf{Trained Qwen3-8B, single-window agent (no context folding or backtracking)}} \\[3pt]
        RL                     & Ablation  & Single Window & 56.0 & 52.1 & 54.6 & 59.5 & 55.0 & \textbf{10K} & \textbf{15} \\
        \midrule
        \multicolumn{10}{@{}l}{\textbf{Trained Qwen3-8B, context folding agent only}} \\[3pt]
        SFT                    & Ablation  & Fold & 40.4 & 34.3 & 45.3 & 51.4 & 42.4 & 31K & 58 \\
        RL                     & Ablation  & Fold & 59.2 & 51.9 & 57.0 & 59.5 & 57.5 & 12K & 23 \\
        SFT + RL               & Ablation  & Fold & 69.0 & 67.8 & 63.6 & 67.6 & 66.3 & 14K & 29 \\
        \midrule
        \multicolumn{10}{@{}l}{\textbf{Trained Qwen3-8B, context folding agent with confidence-based backtracking}} \\[3pt]
        SFT                    & Ablation  & Fold + Backtrack & 59.1 & 57.5 & 59.3 & 64.9 & 59.1 & 73K & 136 \\
        \rowcolor[HTML]{E8F0FB}
        \textbf{\projectname} (SFT + RL) & Proposed & Fold + Backtrack & \textbf{70.2} & \textbf{72.8} & \textbf{64.9} & \textbf{70.3} & \textbf{67.8} & 23K & 48 \\
        \bottomrule
    \end{tabular}%
    }
    \caption{Test accuracy (\%) on LiteraryQA (LitQA), NovelHopQA (NHopQA), NovelQA, and SagaScale (Saga); $n$ is the number of test questions. Tokens are the mean per rollout, and cost is for the full test set. Bold marks the best result among ablation and proposed methods.
\projectname{} is best on every dataset. In the same harness, it nearly matches GLM-5.3 (67.8 vs.\ 69.3 micro) at $1/9$ the cost. It beats GPT-5.5 High in both settings, including full-book context (67.8 vs.\ 56.7), at ${\sim}1/54$ the cost. It is also within 6.5 points of full-context GLM-5.3 at ${\sim}1/26$ the cost.
$^\dagger$Evaluated on up to 100 random questions per dataset; cost is extrapolated to the full test set.}
    \label{tab:main_results}
\end{table*}

\section{Results and Ablations}
\label{sec:results}

We report accuracy on the 4.7K held-out test questions, micro-averaged over the four datasets unless stated otherwise.
Cost is the evaluation cost in USD of running all test rollouts once on an 8$\times$NVIDIA B200 server at an fair market rate of \$64/hr, except for GPT-5.5, for which we report API cost.

\subsection{Comparison with State-of-the-Art Agentic Models}
\label{sec:frontier_comparison}

Table~\ref{tab:main_results} shows that an 8B model trained for agentic Book QA can compete with state-of-the-art (SOTA) models that are far larger and more expensive to run.

\paragraph{Nearly matching a $\sim$100$\times$ larger SOTA agentic model in the same harness.}
Using the same context folding and confidence-based backtracking harness, Table~\ref{tab:main_results} shows that \projectname{} reaches an accuracy of 67.8\% across the four datasets, compared to 69.3\% for the 753B GLM-5.3, at less than $1/9$ the cost and with half the tokens per rollout.
While GLM-5.3 leads on LiteraryQA, NovelQA, and the SagaScale subset, \projectname{} leads on NovelHopQA by 29.3 points. \projectname{} is also the more consistent system, ranging from 64.9\% to 72.8\% across datasets, compared with 43.4\% to 91.9\% for GLM-5.3.
\kmnote*{Shouldn't you refer to results table here? }

\paragraph{Exceeding GPT-5.5 High on agentic Book QA.}
In our agent harness, GPT-5.5 reaches an accuracy of only 13.1\%.
Two failures account for this:
First, GPT-5.5 produces an answer in the required parsable format in only 21.4\% of rollouts.
Second, it never calls the note tool, so each fold leaves it with nothing carried forward.
Even after correcting for output format, it scores only 57.1\%, 10.7 points below \projectname{}.
The strongest proprietary model we tested cannot use a note-based harness that an 8B model can learn through SFT and RL.

\paragraph{Approaching full-context performance at a fraction of the cost.}
\kmnote*{I feel all these numbers in text are somewhat hard to read. Would really like pointing at a table. Then OK to include in text also. although perhaps not necessary and you could save some space. I see you have a table 2 but without the larger models' numbers. Could you put them in that table also? }
As seen in Table~\ref{tab:main_results}, the most accurate system remains GLM-5.3 with the entire book in its 1M-token context, at 74.3\%.
\projectname{} comes within 6.4 points of this result at around $1/26$ of the cost and around $7\times$ fewer tokens per question.
\projectname{} also outperforms GPT-5.5 with the full book in its 232K context window, as GPT-5.5 reaches 56.7\% in this setting, 11.2 points below \projectname{}, at around 54 times the cost.
\projectname{} never holds more than a 20K-token window, yet it answers questions about books that run to millions of tokens.
For comparison, Qwen3-8B with the book truncated to a 40K window only scores 34.9\%.

\paragraph{Training must be co-designed with the test-time harness.}
These comparisons show that a capable model and a well-designed harness are not enough on their own.
Both GPT-5.5 and GLM-5.3 have the context folding and backtracking mechanism available to them with the test-time agent harness, but not the note-taking behavior that makes it useful.
In our evaluation, GPT-5.5 never writes a note, and only 1.2\% of GLM-5.3's attempts that triggered context folding satisfy our note validity criteria. These results suggest that training with the same note tool, prompts, and context layout used at test time helps \projectname use the harness effectively.
Therefore, this experiment is a sign that context management is a behavior that should be learned together with the harness and the task, not one that is achieved through prompting or emerges from scale.
Once effective context management behavior is learned, an 8B model can nearly match a $\sim$100$\times$ larger agent at a fraction of its cost.
\smara*{The results in this section are very strong}

\subsection{Ablation Studies}
\label{sec:ablations}

\begin{figure}[h]
    \centering
    \includegraphics[width=0.88\columnwidth]{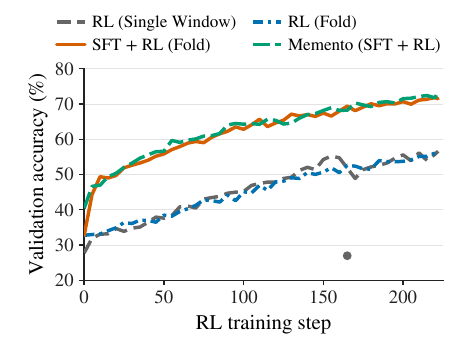}
    \caption{Validation answer accuracy over RL training for the models in Table~\ref{tab:main_results}. The gray dot marks a single outlier evaluation of the RL (Single Window) run, caused by reward hacking that led us to tighten the answer length limit. While SFT + RL and RL only variants start at similar validation accuracy, SFT + RL outperforms RL only by 15\% at the end of training due to more effective compaction behavior instilled by the SFT.} 
    \label{fig:validation_accuracy}
\end{figure}

\begin{figure}[h]
    \centering
    \includegraphics[width=0.88\columnwidth]{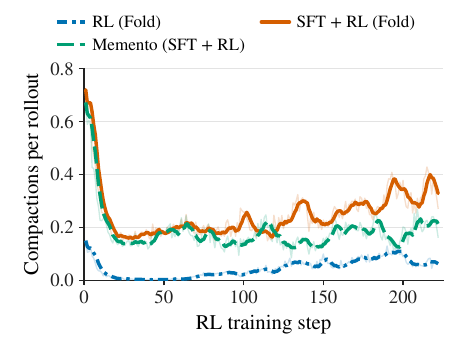}
    \caption{Mean context-window restarts per RL training rollout. Legend names follow Table~\ref{tab:main_results}; training uses compaction into fresh windows. Faint curves show per-step values; bold curves show five-step trailing means. SFT + RL checkpoints retain compaction ratio robustly during the RL training process, while compaction ratio in RL only checkpoints collapses.}
    \label{fig:compaction_ratio}
\end{figure}

To examine the contribution of each element in \projectname{}, we assess the roles of SFT initialization and RL, compare context folding with single-window inference and compaction into a new window, and evaluate confidence based backtracking.

\subsubsection{SFT Initialization and RL}
\label{sec:abl_sft_rl}

Table~\ref{tab:main_results} shows that neither SFT nor RL alone is sufficient to maximize accuracy.
In the context-folding agent, combining SFT with RL improves accuracy by 23.9 points over SFT alone and 8.8 points over RL alone.
The training dynamics of these agents give us insight into why this may be.
As Figure~\ref{fig:validation_accuracy} shows, at the start of RL, the policies with and without SFT initialization have nearly the same validation accuracy, meaning that SFT itself does not improve accuracy.
SFT does, however, change the initial policy's behavior: Figure~\ref{fig:compaction_ratio} shows that the SFT-initialized policy compacts its context more than four times as often as the uninitialized policy before RL.

Figure~\ref{fig:compaction_ratio} also shows that without SFT initialization, RL abandons compaction almost immediately: the compaction rate drops to nearly zero within the first 20 training steps and stays there for the rest of training.
However, with SFT initialization, RL first prunes the excess compactions inherited from SFT, bringing the compaction rate to around 0.15 within about 20 steps, and then slightly increase the compaction rate until the end of the training.
As a result, the validation accuracy for the SFT + RL ablation finishes near 72\% (Figure~\ref{fig:validation_accuracy}), 15\% of the RL only variant, which did not develop effective compaction behavior.

We interpret this as follows.
SFT gives the model the right strategy, such as when to write a note, what to record, and how to resume.
However, imitation alone does not teach the model to carry that strategy out on its own trajectories, so the SFT policy tends to over-compact and accumulate errors across windows.
Since RL learns from the policy's own samples, it reinforces trajectories that execute the strategy well, which the SFT policy only produces occasionally.
Without SFT, RL rarely samples such trajectories, so it has nothing to reinforce and instead learns to avoid the tool.
Both stages are therefore needed: SFT gives RL useful context-management behavior to reinforce, while RL makes that behavior reliable.
\kmnote*{this is a nice paragraph!}

\subsubsection{Context Folding}
\label{sec:abl_fold}
\begin{table}[h]
    \centering
    \small
    \setlength{\tabcolsep}{5pt}
    \begin{tabular}{@{}lccc@{}}
        \toprule
        & Single & + Context & + Back- \\
        Model & window & folding & tracking \\
        \midrule
        Qwen3-8B (untrained)               & 29.6 & 32.6 & 36.9 \\
        SFT                                & 34.4 & 46.6 & 59.1 \\
        \textbf{\projectname{}} (SFT + RL) & 58.6 & 66.9 & \textbf{67.8} \\
        \bottomrule
    \end{tabular}
    \caption{Ablation of our training and inference components. We report test-set accuracy (\%) of the confidence-aware variant of each checkpoint. Rows add training stages (SFT on synthetic data, then RL), and columns add inference-time components cumulatively (context folding, then confidence-aware backtracking). Accuracy improves along both axes: each training stage helps under every inference setup, and each inference component helps for every checkpoint. The best accuracy is shown in bold.}
    \label{tab:stepwise}
\end{table}

First, we compare the model at different training stages with a single context window and with context folding.
Table~\ref{tab:stepwise} shows that context folding improves \projectname{} accuracy by 8.3 points.
However, the untrained model gains only 3.0 points because it rarely reaches the end of its window and usually records nothing useful when it does.
The SFT model gains the most, 12.2 points; it exhausts its first context window in 40.4\% of attempts, making the ability to continue especially useful.
The benefit of folding therefore depends on both the need to continue beyond one window and the ability to preserve useful findings.
For \projectname{}, folding lets the agent carry its findings forward in notes and continue investigating after its window fills.
This extends its effective context well beyond a single window, so it can answer questions whose evidence it could not gather within one.

Then, we utilize an SFT + RL policy trained without confidence to compare compaction into a new window with context folding.
Experiments showed that context folding showed a slight accuracy advantage, scoring 66.27\%, while compaction into a new window scored 65.36\%.  
Context folding is also more token-efficient: because it keeps the turns after the latest note instead of discarding them, it can reduce the need to repeat recent work.
Compared to compacting into a new window, it used 14.3K tokens per rollout versus 15.0K, a 4.7\% saving in token usage.
The model generalizes to context folding without losing accuracy and with fewer tokens, even though it was trained only on compaction into a new window.


\subsubsection{Confidence-Based Backtracking}
\label{sec:abl_backtrack}

Table~\ref{tab:stepwise} shows that confidence-based backtracking helps across different points in training, with the biggest gains seen when the initial answers are weak.
It adds 4.3 points for the untrained model, 12.5 points for the SFT model, and 1.0 point for \projectname{}.

The relatively smaller gain from retries for the RL-trained model reflects a limitation in calibration:
over RL training, the policy first learns to flag its wrong answers as low confidence, but later in training it flags only about 30--40\% of them, and overall agreement rises mainly because it grows confident in more correct answers.
As a result, many of its remaining errors are confident ones that never trigger a retry.
It retries in only 18.9\% of rollouts, and these retries rarely recover yet account for 43\% of its tokens. For a strong policy, backtracking is thus the least cost-effective step and the natural one to drop under a tight budget.
Improving the detection of confident errors is the most direct way to make it more useful, and we leave this to future work.
\section{Conclusion}
\label{sec:conclusion}
We presented \textsc{Memento}, a Qwen3-8B agent that learns context management through compaction notes. SFT on hindsight-annotated teacher trajectories bootstraps note-taking, and RL with a hybrid reward makes it reliable. Although trained only on compaction, the policy transfers to context folding and confidence-based backtracking at test time. On book-level QA, \textsc{Memento} raises accuracy from 36.9\% to 67.8\% with at most 20K tokens in context, nearly matching a $\sim$100$\times$ larger GLM-5.3 agent at 1/9 the cost. Our results suggest that context management should be learned jointly with the task and the harness.

\section*{Limitations}

\paragraph{Task and evaluation scope.} We study only English book-level QA with short free-form answers, graded by an LLM judge. Whether learned note-taking transfers to other long-horizon settings, such as web research, software engineering, or multi-document synthesis, remains untested. GPT-5.5 was evaluated on at most 100 questions per dataset due to API cost. The SagaScale test split contains only 37 questions, so per-dataset results on it are noisy.

\paragraph{Calibration and backtracking.} Late in RL training, the policy flags only about 30--40\% of its wrong answers as low confidence. Many errors therefore never trigger a retry. Backtracking adds just 1.0 point for \textsc{Memento} while accounting for 43\% of its tokens, making it the least cost-effective component for a strong policy.

\paragraph{Training design.} Note validity is enforced through hand-crafted heuristics (frequency, placement, diversity) that may not capture note usefulness. RL uses only compaction into a new window with at most three segments, so longer horizons and the folding and backtracking modes are seen only at test time.

\makeatletter
\ifacl@finalcopy
\section*{Acknowledgments}
Acknowledgments and funding information
\fi
\makeatother

\bibliography{custom}

@inproceedings{yao2023react,
    title     = {{ReAct}: Synergizing Reasoning and Acting in Language Models},
    author    = {Yao, Shunyu and Zhao, Jeffrey and Yu, Dian and Du, Nan and Shafran, Izhak and Narasimhan, Karthik and Cao, Yuan},
    booktitle = {International Conference on Learning Representations (ICLR)},
    year      = {2023},
    url       = {https://openreview.net/forum?id=WE_vluYUL-X}
}

@article{liu2024lost,
    title     = {Lost in the Middle: How Language Models Use Long Contexts},
    author    = {Liu, Nelson F. and Lin, Kevin and Hewitt, John and Paranjape, Ashwin and Bevilacqua, Michele and Petroni, Fabio and Liang, Percy},
    journal   = {Transactions of the Association for Computational Linguistics},
    year      = {2024},
    volume    = {12},
    pages     = {157--173},
    publisher = {MIT Press},
    doi       = {10.1162/tacl_a_00638},
    url       = {https://aclanthology.org/2024.tacl-1.9/}
}

@inproceedings{du2025context,
    title     = {Context Length Alone Hurts {LLM} Performance Despite Perfect Retrieval},
    author    = {Du, Yufeng and Tian, Minyang and Ronanki, Srikanth and Rongali, Subendhu and Bodapati, Sravan and Galstyan, Aram and Wells, Azton and Schwartz, Roy and Huerta, Eliu A. and Peng, Hao},
    booktitle = {Findings of the Association for Computational Linguistics: EMNLP 2025},
    year      = {2025},
    pages     = {23281--23298},
    publisher = {Association for Computational Linguistics},
    address   = {Suzhou, China},
    doi       = {10.18653/v1/2025.findings-emnlp.1264},
    url       = {https://aclanthology.org/2025.findings-emnlp.1264/}
}

@article{packer2023memgpt,
    title     = {{MemGPT}: Towards {LLMs} as Operating Systems},
    author    = {Packer, Charles and Wooders, Sarah and Lin, Kevin and Fang, Vivian and Patil, Shishir G. and Stoica, Ion and Gonzalez, Joseph E.},
    journal   = {arXiv preprint arXiv:2310.08560},
    year      = {2023},
    doi       = {10.48550/arXiv.2310.08560},
    url       = {https://arxiv.org/abs/2310.08560}
}

@article{wu2025resum,
    title     = {{ReSum}: Unlocking Long-Horizon Search Intelligence via Context Summarization},
    author    = {Wu, Xixi and Li, Kuan and Zhao, Yida and Zhang, Liwen and Ou, Litu and Yin, Huifeng and Zhang, Zhongwang and Yu, Xinmiao and Zhang, Dingchu and Jiang, Yong and Xie, Pengjun and Huang, Fei and Cheng, Minhao and Wang, Shuai and Cheng, Hong and Zhou, Jingren},
    journal   = {arXiv preprint arXiv:2509.13313},
    year      = {2025},
    doi       = {10.48550/arXiv.2509.13313},
    url       = {https://arxiv.org/abs/2509.13313}
}

@inproceedings{zhou2025mem1,
    title     = {{MEM1}: Learning to Synergize Memory and Reasoning for Efficient Long-Horizon Agents},
    author    = {Zhou, Zijian and Qu, Ao and Wu, Zhaoxuan and Kim, Sunghwan and Prakash, Alok and Rus, Daniela and Low, Bryan Kian Hsiang and Liang, Paul Pu},
    booktitle = {International Conference on Learning Representations (ICLR)},
    year      = {2026},
    pages     = {58413--58438},
    url       = {https://proceedings.iclr.cc/paper_files/paper/2026/hash/5fc8b3bdfbb9167b5144df5d3fae4616-Abstract-Conference.html}
}

@inproceedings{yu2025memagent,
    title     = {{MemAgent}: Reshaping Long-Context {LLM} with Multi-Conv {RL}-based Memory Agent},
    author    = {Yu, Hongli and Chen, Tinghong and Feng, Jiangtao and Chen, Jiangjie and Dai, Weinan and Yu, Qiying and Zhang, Ya-Qin and Ma, Wei-Ying and Liu, Jingjing and Wang, Mingxuan and Zhou, Hao},
    booktitle = {International Conference on Learning Representations (ICLR)},
    year      = {2026},
    pages     = {39458--39486},
    url       = {https://proceedings.iclr.cc/paper_files/paper/2026/hash/4264ee4376776907c0b87ed70b959585-Abstract-Conference.html}
}

@inproceedings{lu2025supo,
    title     = {Beyond the Context Window: Scaling Agentic {RL} via End-to-end Optimized Context Compression},
    author    = {Lu, Miao and Sun, Weiwei and Du, Weihua and Ling, Zhan and Yao, Xuesong and Liu, Kang and Chen, Jiecao},
    year      = {2026},
    doi       = {10.18653/v1/2026.acl-long.966},
    url       = {https://aclanthology.org/2026.acl-long.966/},
    booktitle = {Proceedings of the 64th Annual Meeting of the Association for Computational Linguistics (Volume 1: Long Papers)},
    pages     = {21074--21125},
    publisher = {Association for Computational Linguistics},
    address   = {San Diego, California, United States}
}

@inproceedings{sun2025folding,
    title     = {Scaling Long-Horizon Agent via Context-Folding},
    author    = {Sun, Weiwei and Lu, Miao and Ling, Zhan and Liu, Kang and Yao, Xuesong and Yang, Yiming and Chen, Jiecao},
    year      = {2026},
    url       = {https://proceedings.mlr.press/v306/sun26x.html},
    booktitle = {Proceedings of the 43rd International Conference on Machine Learning},
    volume    = {306},
    series    = {Proceedings of Machine Learning Research},
    pages     = {116894--116910},
    publisher = {PMLR}
}

@inproceedings{ye2025agentfold,
    title     = {{AgentFold}: Long-Horizon Web Agents with Proactive Context Folding},
    author    = {Ye, Rui and Zhang, Zhongwang and Li, Kuan and Yin, Huifeng and Tao, Zhengwei and Zhao, Yida and Su, Liangcai and Zhang, Liwen and Yin, Wenbiao and Qiao, Zile and Wang, Xinyu and Xie, Pengjun and Huang, Fei and Zhou, Jingren and Chen, Siheng and Jiang, Yong},
    booktitle = {International Conference on Learning Representations (ICLR)},
    year      = {2026},
    pages     = {154754--154776},
    url       = {https://proceedings.iclr.cc/paper_files/paper/2026/hash/fa948624dfde013671e72c1a7ca4aebc-Abstract-Conference.html}
}

@article{shao2025foldact,
    title     = {{FoldAct}: Efficient and Stable Context Folding for Long-Horizon Search Agents},
    author    = {Shao, Jiaqi and Miao, Yufeng and Zhang, Wei and Luo, Bing},
    journal   = {arXiv preprint arXiv:2512.22733},
    year      = {2025},
    doi       = {10.48550/arXiv.2512.22733},
    url       = {https://arxiv.org/abs/2512.22733}
}

@inproceedings{bonomo2025literaryqa,
    title     = {{LiteraryQA}: Towards Effective Evaluation of Long-document Narrative {QA}},
    author    = {Bonomo, Tommaso and Gioffr{\'e}, Luca and Navigli, Roberto},
    booktitle = {Proceedings of the 2025 Conference on Empirical Methods in Natural Language Processing},
    year      = {2025},
    pages     = {34086--34107},
    publisher = {Association for Computational Linguistics},
    address   = {Suzhou, China},
    doi       = {10.18653/v1/2025.emnlp-main.1729},
    url       = {https://aclanthology.org/2025.emnlp-main.1729/}
}

@inproceedings{gupta2025novelhopqa,
    title     = {{NovelHopQA}: Diagnosing Multi-Hop Reasoning Failures in Long Narrative Contexts},
    author    = {Gupta, Abhay and Lu, Michael and Zhu, Kevin and O'Brien, Sean and Sharma, Vasu},
    booktitle = {Proceedings of the 2025 Conference on Empirical Methods in Natural Language Processing},
    year      = {2025},
    pages     = {26134--26151},
    publisher = {Association for Computational Linguistics},
    address   = {Suzhou, China},
    doi       = {10.18653/v1/2025.emnlp-main.1328},
    url       = {https://aclanthology.org/2025.emnlp-main.1328/}
}

@article{du2026sagascale,
    title     = {{SagaScale}: A Realistic, Scalable, and High-Quality Long-Context Benchmark Built from Full-Length Novels},
    author    = {Du, Guancheng and Hu, Yong and Wang, Wenqing and Yang, Yaming and Gao, Jiaheng},
    journal   = {arXiv preprint arXiv:2601.09723},
    year      = {2026},
    doi       = {10.48550/arXiv.2601.09723},
    url       = {https://arxiv.org/abs/2601.09723}
}

@article{chen2023memwalker,
    title     = {Walking Down the Memory Maze: Beyond Context Limit through Interactive Reading},
    author    = {Chen, Howard and Pasunuru, Ramakanth and Weston, Jason and Celikyilmaz, Asli},
    journal   = {arXiv preprint arXiv:2310.05029},
    year      = {2023},
    doi       = {10.48550/arXiv.2310.05029},
    url       = {https://arxiv.org/abs/2310.05029}
}

@inproceedings{lee2024readagent,
    title     = {A Human-Inspired Reading Agent with Gist Memory of Very Long Contexts},
    author    = {Lee, Kuang-Huei and Chen, Xinyun and Furuta, Hiroki and Canny, John and Fischer, Ian},
    booktitle = {Proceedings of the 41st International Conference on Machine Learning},
    year      = {2024},
    volume    = {235},
    series    = {Proceedings of Machine Learning Research},
    pages     = {26396--26415},
    publisher = {PMLR},
    url       = {https://proceedings.mlr.press/v235/lee24c.html}
}

@inproceedings{li2024graphreader,
    title     = {{GraphReader}: Building Graph-based Agent to Enhance Long-Context Abilities of Large Language Models},
    author    = {Li, Shilong and He, Yancheng and Guo, Hangyu and Bu, Xingyuan and Bai, Ge and Liu, Jie and Liu, Jiaheng and Qu, Xingwei and Li, Yangguang and Ouyang, Wanli and Su, Wenbo and Zheng, Bo},
    booktitle = {Findings of the Association for Computational Linguistics: EMNLP 2024},
    year      = {2024},
    pages     = {12758--12786},
    publisher = {Association for Computational Linguistics},
    address   = {Miami, Florida, USA},
    doi       = {10.18653/v1/2024.findings-emnlp.746},
    url       = {https://aclanthology.org/2024.findings-emnlp.746/}
}

@article{shao2024deepseekmath,
    title     = {{DeepSeekMath}: Pushing the Limits of Mathematical Reasoning in Open Language Models},
    author    = {Shao, Zhihong and Wang, Peiyi and Zhu, Qihao and Xu, Runxin and Song, Junxiao and Bi, Xiao and Zhang, Haowei and Zhang, Mingchuan and Li, Y. K. and Wu, Y. and Guo, Daya},
    journal   = {arXiv preprint arXiv:2402.03300},
    year      = {2024},
    doi       = {10.48550/arXiv.2402.03300},
    url       = {https://arxiv.org/abs/2402.03300}
}

@inproceedings{wang2023selfconsistency,
    title     = {Self-Consistency Improves Chain of Thought Reasoning in Language Models},
    author    = {Wang, Xuezhi and Wei, Jason and Schuurmans, Dale and Le, Quoc and Chi, Ed H. and Narang, Sharan and Chowdhery, Aakanksha and Zhou, Denny},
    booktitle = {International Conference on Learning Representations (ICLR)},
    year      = {2023},
    url       = {https://openreview.net/forum?id=1PL1NIMMrw}
}

@inproceedings{yao2023tot,
    title     = {Tree of Thoughts: Deliberate Problem Solving with Large Language Models},
    author    = {Yao, Shunyu and Yu, Dian and Zhao, Jeffrey and Shafran, Izhak and Griffiths, Thomas L. and Cao, Yuan and Narasimhan, Karthik},
    booktitle = {Advances in Neural Information Processing Systems},
    year      = {2023},
    volume    = {36},
    pages     = {11809--11822},
    doi       = {10.52202/075280-0517},
    url       = {https://proceedings.neurips.cc/paper_files/paper/2023/hash/271db9922b8d1f4dd7aaef84ed5ac703-Abstract-Conference.html}
}

@inproceedings{shinn2023reflexion,
    title     = {Reflexion: Language Agents with Verbal Reinforcement Learning},
    author    = {Shinn, Noah and Cassano, Federico and Gopinath, Ashwin and Narasimhan, Karthik and Yao, Shunyu},
    booktitle = {Advances in Neural Information Processing Systems},
    year      = {2023},
    volume    = {36},
    pages     = {8634--8652},
    doi       = {10.52202/075280-0377},
    url       = {https://proceedings.neurips.cc/paper_files/paper/2023/hash/1b44b878bb782e6954cd888628510e90-Abstract-Conference.html}
}

@inproceedings{tian2023justask,
    title     = {Just Ask for Calibration: Strategies for Eliciting Calibrated Confidence Scores from Language Models Fine-Tuned with Human Feedback},
    author    = {Tian, Katherine and Mitchell, Eric and Zhou, Allan and Sharma, Archit and Rafailov, Rafael and Yao, Huaxiu and Finn, Chelsea and Manning, Christopher D.},
    booktitle = {Proceedings of the 2023 Conference on Empirical Methods in Natural Language Processing},
    year      = {2023},
    pages     = {5433--5442},
    publisher = {Association for Computational Linguistics},
    address   = {Singapore},
    doi       = {10.18653/v1/2023.emnlp-main.330},
    url       = {https://aclanthology.org/2023.emnlp-main.330/}
}

@inproceedings{damani2025rlcr,
    title     = {Beyond Binary Rewards: Training {LMs} to Reason About Their Uncertainty},
    author    = {Damani, Mehul and Puri, Isha and Slocum, Stewart and Shenfeld, Idan and Choshen, Leshem and Kim, Yoon and Andreas, Jacob},
    booktitle = {International Conference on Learning Representations (ICLR)},
    year      = {2026},
    pages     = {59456--59493},
    url       = {https://proceedings.iclr.cc/paper_files/paper/2026/hash/615675cc6e94ddb1a783904fb178b5f6-Abstract-Conference.html}
}

@misc{qwen3technicalreport,
    title         = {{Qwen3} Technical Report},
    author        = {{Qwen Team}},
    year          = {2025},
    eprint        = {2505.09388},
    archivePrefix = {arXiv},
    primaryClass  = {cs.CL},
    doi           = {10.48550/arXiv.2505.09388},
    url           = {https://arxiv.org/abs/2505.09388}
}

@misc{zai2026glm53,
    title        = {{GLM-5.3}: Frontier Coding with Emergent Cyber Capabilities},
    author       = {{Z.ai}},
    year         = {2026},
    month        = aug,
    howpublished = {Model release announcement},
    url          = {https://z.ai/blog/glm-5.3}
}

@inproceedings{xiao2024cpack,
    title     = {{C-Pack}: Packed Resources for General {Chinese} Embeddings},
    author    = {Xiao, Shitao and Liu, Zheng and Zhang, Peitian and Muennighoff, Niklas and Lian, Defu and Nie, Jian-Yun},
    booktitle = {Proceedings of the 47th International ACM SIGIR Conference on Research and Development in Information Retrieval},
    year      = {2024},
    pages     = {641--649},
    publisher = {Association for Computing Machinery},
    address   = {Washington, DC, USA},
    doi       = {10.1145/3626772.3657878},
    url       = {https://doi.org/10.1145/3626772.3657878}
}

@inproceedings{wang2025novelqa,
  title     = {{NovelQA}: Benchmarking Question Answering on Documents Exceeding 200{K} Tokens},
  author    = {Wang, Cunxiang and Ning, Ruoxi and Pan, Boqi and Wu, Tonghui and Guo, Qipeng and Deng, Cheng and Bao, Guangsheng and Hu, Xiangkun and Zhang, Zheng and Wang, Qian and Zhang, Yue},
  booktitle = {International Conference on Learning Representations},
  year      = {2025},
  url       = {https://proceedings.iclr.cc/paper_files/paper/2025/hash/3adb85a348a18cdd74ce99fbbab20301-Abstract-Conference.html},
  pages     = {23444--23466}
}

@inproceedings{kwa2025measuring,
  title     = {Measuring {AI} Ability to Complete Long Software Tasks},
  author    = {Kwa, Thomas and West, Ben and Becker, Joel and Deng, Amy and Garcia, Katharyn and Hasin, Max and Jawhar, Sami and Kinniment, Megan and Rush, Nate and Von Arx, Sydney and Bloom, Ryan and Broadley, Thomas and Du, Haoxing and Goodrich, Brian and Jurkovic, Nikola and Miles, Luke Harold and Nix, Seraphina and Lin, Tao and Painter, Chris and Parikh, Neev and Rein, David and Sato, Lucas Jun Koba and Wijk, Hjalmar and Ziegler, Daniel M. and Barnes, Elizabeth and Chan, Lawrence},
  booktitle = {Advances in Neural Information Processing Systems},
  year      = {2025},
  url       = {https://proceedings.neurips.cc/paper_files/paper/2025/hash/85069585133c4c168c865e65d72e9775-Abstract-Conference.html},
  volume    = {38},
  pages     = {92213--92266},
  doi       = {10.52202/085713-3086}
}

@article{wei2024simpleqa,
  title   = {Measuring Short-form Factuality in Large Language Models},
  author  = {Wei, Jason and {Nguyen Karina} and Chung, Hyung Won and Jiao, Yunxin Joy and Papay, Spencer and Glaese, Amelia and Schulman, John and Fedus, William},
  journal = {arXiv preprint arXiv:2411.04368},
  year    = {2024},
  doi     = {10.48550/arXiv.2411.04368},
  url     = {https://arxiv.org/abs/2411.04368}
}

@article{wei2025browsecomp,
  title   = {{BrowseComp}: A Simple Yet Challenging Benchmark for Browsing Agents},
  author  = {Wei, Jason and Sun, Zhiqing and Papay, Spencer and McKinney, Scott and Han, Jeffrey and Fulford, Isa and Chung, Hyung Won and Passos, Alex Tachard and Fedus, William and Glaese, Amelia},
  journal = {arXiv preprint arXiv:2504.12516},
  year    = {2025},
  doi     = {10.48550/arXiv.2504.12516},
  url     = {https://arxiv.org/abs/2504.12516}
}

@inproceedings{mialon2024gaia,
  title     = {{GAIA}: a benchmark for General {AI} Assistants},
  author    = {Mialon, Gr{\'e}goire and Fourrier, Cl{\'e}mentine and Swift, Craig and Wolf, Thomas and LeCun, Yann and Scialom, Thomas},
  booktitle = {International Conference on Learning Representations},
  year      = {2024},
  url       = {https://proceedings.iclr.cc/paper_files/paper/2024/hash/25ae35b5b1738d80f1f03a8713e405ec-Abstract-Conference.html},
  pages     = {9025--9049}
}

@inproceedings{yang2018hotpotqa,
  title     = {{HotpotQA}: A Dataset for Diverse, Explainable Multi-hop Question Answering},
  author    = {Yang, Zhilin and Qi, Peng and Zhang, Saizheng and Bengio, Yoshua and Cohen, William W. and Salakhutdinov, Ruslan and Manning, Christopher D.},
  booktitle = {Proceedings of the 2018 Conference on Empirical Methods in Natural Language Processing},
  pages     = {2369--2380},
  year      = {2018},
  address   = {Brussels, Belgium},
  publisher = {Association for Computational Linguistics},
  doi       = {10.18653/v1/D18-1259},
  url       = {https://aclanthology.org/D18-1259/}
}

@inproceedings{ho2020constructing,
  title     = {Constructing A Multi-hop {QA} Dataset for Comprehensive Evaluation of Reasoning Steps},
  author    = {Ho, Xanh and Duong Nguyen, Anh-Khoa and Sugawara, Saku and Aizawa, Akiko},
  booktitle = {Proceedings of the 28th International Conference on Computational Linguistics},
  pages     = {6609--6625},
  year      = {2020},
  address   = {Barcelona, Spain (Online)},
  publisher = {International Committee on Computational Linguistics},
  doi       = {10.18653/v1/2020.coling-main.580},
  url       = {https://aclanthology.org/2020.coling-main.580/}
}

@article{trivedi2022musique,
  title     = {{MuSiQue}: Multihop Questions via Single-hop Question Composition},
  author    = {Trivedi, Harsh and Balasubramanian, Niranjan and Khot, Tushar and Sabharwal, Ashish},
  journal   = {Transactions of the Association for Computational Linguistics},
  volume    = {10},
  pages     = {539--554},
  year      = {2022},
  publisher = {MIT Press},
  doi       = {10.1162/tacl_a_00475},
  url       = {https://aclanthology.org/2022.tacl-1.31/}
}

@misc{aa2026glm53,
   author       = {{Artificial Analysis}},
   title        = {{GLM-5.3 (max)}: Intelligence, Performance \& Price Analysis},
   year         = {2026},
   howpublished = {\url{https://artificialanalysis.ai/models/glm-5-3}},
   note         = {Accessed: 2026-10-12}
 }

@misc{ha2026memguard,
  title         = {{MemGuard}: Preventing Memory Contamination in
                   Long-Term Memory-Augmented Large Language Models},
  author        = {Ha, Hyeonjeong and Kim, Jeonghwan and Qian, Cheng
                   and Liu, Jiayu and Campbell, William M. and Wu, Yue
                   and Zhang, Yuji and McKeown, Kathleen
                   and Hakkani-T{\"u}r, Dilek and Ji, Heng},
  year          = {2026},
  eprint        = {2605.28009},
  archivePrefix = {arXiv},
  primaryClass  = {cs.CL},
  url           = {https://arxiv.org/abs/2605.28009}
}

\clearpage
\appendix

\section{Training Settings}
\label{sec:training_settings}

Tables~\ref{tab:sft_hyperparameters}--\ref{tab:generation_hyperparameters} give the settings and library versions used for the reported checkpoints and evaluations.
Settings are shared across conditions unless an exception is stated.

%
\begin{table}[H]
    \centering
    \small
    \setlength{\tabcolsep}{3pt}
    \begin{tabular}{@{}p{0.48\columnwidth}p{0.47\columnwidth}@{}}
        \toprule
        Hyperparameter & Value \\
        \midrule
        Initial model & Qwen3-8B \\
        Optimizer & AdamW \\
        Peak / minimum learning rate & $2\times10^{-5}$ / $4\times10^{-6}$ \\
        Learning-rate schedule & Single cosine decay \\
        Warmup & $0.1$ epoch \\
        Weight decay & $0.01$ \\
        Adam momentum $(\beta_1,\beta_2)$ & $(0.9,0.98)$ \\
        Gradient-clipping threshold & $5.0$ \\
        Global batch size & 64 segments \\
        Microbatch size per GPU & 2 segments \\
        Gradient accumulation & 4 steps \\
        Training epochs & 2 \\
        Maximum sequence length & 18,432 tokens \\
        Loss-bearing tokens & Assistant tokens only \\
        Precision / parallelism & BF16 / FSDP, 8 GPUs \\
        Activation checkpointing & Enabled \\
        \midrule
        \multicolumn{2}{@{}l}{\emph{Libraries}} \\
        PyTorch & 2.10.0+cu129 \\
        HF Transformers & 5.4.0 \\
        HF Hub & 1.24.0 \\
        HF Tokenizers & 0.22.2 \\
        FlashAttention 4 & 4.0.0b22.dev2 \\
        \midrule
        \multicolumn{2}{@{}l}{\emph{Compute resources}} \\
        GPUs & 8$\times$ NVIDIA B200 \\
        Estimated total GPU hours & $\sim$25 per experiment \\
        \bottomrule
    \end{tabular}
    \caption{SFT hyperparameters, shared by the variants with and without confidence. Sequences exceeding the limit are truncated from the left, and an assistant turn cut by truncation is excluded from the loss. We use the final checkpoint after two epochs.}
    \label{tab:sft_hyperparameters}
\end{table}

\newpage 

\begin{table}[H]
    \centering
    \small
    \setlength{\tabcolsep}{3pt}
    \begin{tabular}{@{}p{0.48\columnwidth}p{0.47\columnwidth}@{}}
        \toprule
        Hyperparameter & Value \\
        \midrule
        Advantage estimator & GRPO \\
        RL rollout batch size & 256 questions \\
        Policy minibatch size & 64 questions \\
        GRPO group size & 16 \\
        Policy update epochs & 1 per rollout batch \\
        Max context window & 4K prompt / 12K output \\
        Max segments per rollout & 3 \\
        Training duration & 3 epochs; 222 rollout steps \\
        Optimizer & AdamW \\
        Learning rate & $10^{-6}$, constant \\
        Learning-rate warmup & None \\
        Weight decay & $0.01$ \\
        Adam momentum $(\beta_1,\beta_2)$ & $(0.9,0.98)$ \\
        GRPO clipping thresholds & $[0.8,1.28]$ \\
        KL-loss coefficient & $10^{-4}$ \\
        Entropy coefficient & $0$ \\
        Loss aggregation & Mean over generated tokens \\
        Gradient-clipping threshold & $2.0$ \\
        Validation interval & Every 5 rollout steps \\
        \midrule
        \multicolumn{2}{@{}l}{\emph{Libraries}} \\
        verl (modified) & 0.8.0.dev0 \\
        PyTorch & 2.10.0+cu129 \\
        HF Transformers & 5.4.0 \\
        HF Hub & 1.24.0 \\
        HF Tokenizers & 0.22.2 \\
        FlashAttention 4 & 4.0.0b22.dev2 \\
        vLLM & 0.18.0 \\
        \midrule
        \multicolumn{2}{@{}l}{\emph{Compute resources}} \\
        GPUs & 8$\times$ NVIDIA B200 \\
        Estimated total GPU hours & $\sim$1{,}000 per experiment \\
        \bottomrule
    \end{tabular}
    \caption{GRPO hyperparameters, shared by the four RL conditions. The policy and fixed KL reference start from the corresponding SFT checkpoint, or from Qwen3-8B for the RL-only baselines. The confidence reward weight is $\lambda=0.10$ through step 160 and $0.25$ thereafter, and variants without confidence use $\lambda=0$. Reported RL results use step 222.}
    \label{tab:grpo_hyperparameters}
\end{table}

\newpage

\newpage
\begin{table}[!htbp]
    \centering
    \small
    \setlength{\tabcolsep}{3pt}
    \begin{tabular}{@{}p{0.48\columnwidth}p{0.47\columnwidth}@{}}
        \toprule
        Hyperparameter & Value \\
        \midrule
        \multicolumn{2}{@{}l}{\emph{Policy sampling}} \\
        Qwen temperature / top-$p$ & $1.0$ / $1.0$ \\
        Qwen top-$k$ & Off; untrained test baseline: $20$ \\
        Validation / test samples & 1 per question, before retries \\
        GLM test temperature / top-$p$ & $1.0$ / $0.95$ \\
        GLM test reasoning effort & High \\
        \midrule
        \multicolumn{2}{@{}l}{\emph{Context and generation budgets (tokens)}} \\
        \shortstack[l]{Compaction agent \\prompt / response soft limit} & 4,096 / 12,288 \\
        \shortstack[l]{Compaction agent \\ hard context limit} & 20,480 \\
        Max segment & 3, in training and test \\
        Maximum folds & 7 per attempt \\
        \shortstack[l]{GLM full-book\\input / output limit} & 900,000 / 100,000 \\
        \shortstack[l]{GPT-5.5 full-book\\input / output limit} & 204,800 / 32,768 \\

        \midrule
        \multicolumn{2}{@{}l}{\emph{Agent limits}} \\
        Assistant / user--tool turns & 256 / 256 per session \\
        Parallel tool calls & 8 per turn \\
        Tool-response limit & 12,000 characters \\
        Retained notes & \shortstack[l]{Latest 5; at most\\400 tokens per note} \\
        Valid note length & 40--400 policy tokens \\
        \midrule
        \multicolumn{2}{@{}l}{\emph{Confidence search at test time}} \\
        Maximum answer attempts & 5 (4 retries) \\
        Retry threshold & $0.75$ for RL with confidence; $0.5$ otherwise \\
        Retry 1 / retry 2 note anchor & $\lfloor n/2\rfloor$ / $\min(\lfloor n/2\rfloor,1)$ \\
        Retries 3 and 4 & Fresh attempts \\
        Early stopping & Confidence $>$ threshold \\
        Final selection & Highest confidence \\
        \midrule
        \multicolumn{2}{@{}l}{\emph{Answer judges}} \\
        GRPO judge & Qwen3-14B \\
        Test judge & Qwen3-235B-A22B-\allowbreak Instruct-2507-FP8 \\
        Temperature / top-$p$ / top-$k$ & $0.7$ / $0.8$ / $20$ \\
        Generation limit / thinking & 3,072 tokens / off \\
        \bottomrule
    \end{tabular}
    \caption{Inference and generation hyperparameters. Qwen sampling applies to GRPO, validation, and test runs. Response budgets include tool observations, except for per-turn generation caps; the folding segment budget is a soft limit. Tool outputs are truncated from the right; full-book inputs retain the beginning. In the retry schedule, $n$ counts note calls accumulated at that retry. Notes and retries do not apply to the no-compaction baseline.}
    \label{tab:generation_hyperparameters}
\end{table}

\clearpage

\raggedbottom
\section{Sample Trajectory}
\label{sec:trajectory_examples}

We show an example of a Qwen3-8B trajectory with succesful context folding and backtracking.
The first answer, New York, has low confidence and is judged incorrect.
Backtracking to the first saved note leads to Paris, with very high confidence, which the evaluator marks correct.
We reproduce the agent input and every assistant turn, including all recorded reasoning, tool calls, notes, and answer attempts.
The agent system and user prompts can be seen in Appendix~\ref{sec:student_prompts}.
Tool outputs are omitted for brevity, and restored context is identified by transition labels.
The transcript includes the model's original mistakes.

\definecolor{trajectoryturn}{HTML}{426982}
\definecolor{trajectoryturnbg}{HTML}{EDF3F7}
\definecolor{trajectorysaved}{HTML}{317C71}
\definecolor{trajectorysavedbg}{HTML}{E7F2EE}
\definecolor{trajectorycontinuation}{HTML}{79668B}
\definecolor{trajectorycontinuationbg}{HTML}{F1EDF5}
\definecolor{trajectoryinput}{HTML}{566771}
\definecolor{trajectoryinputbg}{HTML}{F4F6F7}

\newcommand{\trajectorykey}[2]{%
  {\color{#1}\rule{6pt}{6pt}\enspace #2}}
\newenvironment{trajectorytext}[1]{%
  \par\addvspace{6pt}%
  \normalfont\normalsize\raggedright
  \setlength{\parindent}{0pt}\setlength{\parskip}{0pt}%
  \setlength{\leftskip}{6pt}\setlength{\rightskip}{6pt plus 1fil}%
  \colorlet{appendixaccent}{#1}%
  \appendixboxstart[#1bg]%
}{\par\appendixboxend\addvspace{12pt}}
\newenvironment{trajectorynote}[1]{%
  \par\addvspace{10pt}%
  \colorlet{appendixaccent}{trajectorysaved}%
  \appendixboxstart[trajectorysavedbg]%
  \excerptlabel{Saved note #1}%
}{\par\appendixboxend\addvspace{8pt}}

\noindent Colors follow Figure~\ref{fig:compaction_methods}:\par
\noindent\trajectorykey{trajectoryturn}{Original turns}\quad
\trajectorykey{trajectorysaved}{Saved notes}\par
\noindent\trajectorykey{trajectorycontinuation}{Continuation after folding or backtracking}\par
\noindent\trajectorykey{trajectoryinput}{Agent input}\par

\excerptlabel{LiteraryQA: Beyond the Rocks}
\begin{trajectorytext}{trajectoryinput}
\excerptlabel{Agent Input}
Book title: Beyond the Rocks\par
Question: Where does Josiah travel to after his health improves?\par
\end{trajectorytext}

\begin{trajectorytext}{trajectoryturn}
\excerptlabel{Initial context; attempt 1}
\excerptlabel{Assistant turn 1}
\textless{}think\textgreater{}\par\nopagebreak
Need to find the book \textquotedbl{}Beyond the Rocks\textquotedbl{} and the character Josiah. Search.\par
\textless{}/\allowbreak{}think\textgreater{}\par
\smallskip
I need to find the book \textquotedbl{}Beyond the Rocks\textquotedbl{} and locate information about Josiah\textquotesingle{}s health and travels after he recovers.\par
\textless{}tool\_\allowbreak{}call\textgreater{}\par\nopagebreak
\{\textquotedbl{}name\textquotedbl{}: \textquotedbl{}ripgrep\_\allowbreak{}search\textquotedbl{},\allowbreak{} \textquotedbl{}arguments\textquotedbl{}: \{\textquotedbl{}pattern\textquotedbl{}: \textquotedbl{}Josiah\textquotedbl{},\allowbreak{} \textquotedbl{}number\_\allowbreak{}of\_\allowbreak{}results\textquotedbl{}: 15\}\}\par
\textless{}/\allowbreak{}tool\_\allowbreak{}call\textgreater{}\par
\textless{}tool\_\allowbreak{}call\textgreater{}\par\nopagebreak
\{\textquotedbl{}name\textquotedbl{}: \textquotedbl{}semantic\_\allowbreak{}search\textquotedbl{},\allowbreak{} \textquotedbl{}arguments\textquotedbl{}: \{\textquotedbl{}query\textquotedbl{}: \textquotedbl{}Josiah travels to a place after his health improves\textquotedbl{},\allowbreak{} \textquotedbl{}number\_\allowbreak{}of\_\allowbreak{}results\textquotedbl{}: 10\}\}\par
\textless{}/\allowbreak{}tool\_\allowbreak{}call\textgreater{}\par
\end{trajectorytext}

\begin{trajectorytext}{trajectoryturn}
\excerptlabel{Assistant turn 2}
\textless{}think\textgreater{}\par\nopagebreak
The question: \textquotedbl{}Where does Josiah travel to after his health improves?\textquotedbl{} Let me read around line 32-35; it mentions winter at Hyères and then he travels. Likely Spain? Let me read the text after he improves.\par
\textless{}/\allowbreak{}think\textgreater{}\par
\smallskip
The early chapters mention Hyères and Paris. Let me read the context around the illness and recovery.\par
\textless{}tool\_\allowbreak{}call\textgreater{}\par\nopagebreak
\{\textquotedbl{}name\textquotedbl{}: \textquotedbl{}view\textquotedbl{},\allowbreak{} \textquotedbl{}arguments\textquotedbl{}: \{\textquotedbl{}line\_\allowbreak{}start\textquotedbl{}: 32,\allowbreak{} \textquotedbl{}line\_\allowbreak{}end\textquotedbl{}: 40\}\}\par
\textless{}/\allowbreak{}tool\_\allowbreak{}call\textgreater{}\par
\textless{}tool\_\allowbreak{}call\textgreater{}\par\nopagebreak
\{\textquotedbl{}name\textquotedbl{}: \textquotedbl{}ripgrep\_\allowbreak{}search\textquotedbl{},\allowbreak{} \textquotedbl{}arguments\textquotedbl{}: \{\textquotedbl{}number\_\allowbreak{}of\_\allowbreak{}results\textquotedbl{}: 10,\allowbreak{} \textquotedbl{}pattern\textquotedbl{}: \textquotedbl{}health improved|\allowbreak{}improved\textquotedbl{},\allowbreak{} \textquotedbl{}results\_\allowbreak{}starting\_\allowbreak{}index\textquotedbl{}: 8\}\}\par
\textless{}/\allowbreak{}tool\_\allowbreak{}call\textgreater{}\par
\end{trajectorytext}

\begin{trajectorytext}{trajectoryturn}
\excerptlabel{Assistant turn 3}
\textless{}think\textgreater{}\par\nopagebreak
I\textquotesingle{}ve just established what this book is and found the illness/\allowbreak{}recovery storyline,\allowbreak{} with the key names. If I lose track of these line numbers I\textquotesingle{}ll have to redo all the searching,\allowbreak{} so let me record what I\textquotesingle{}ve got so far.\par
\textless{}/\allowbreak{}think\textgreater{}\par
\smallskip
I\textquotesingle{}ve found the key plot thread; let me save it before I dig into the aftermath.\par
\begin{trajectorynote}{1}
\textless{}tool\_\allowbreak{}call\textgreater{}\par\nopagebreak
\{\textquotedbl{}name\textquotedbl{}: \textquotedbl{}create\_\allowbreak{}notes\textquotedbl{},\allowbreak{} \textquotedbl{}arguments\textquotedbl{}: \{\textquotedbl{}summary\textquotedbl{}: \textquotedbl{}The book here is \textbackslash{}\textquotedbl{}Beyond the Rocks\textbackslash{}\textquotedbl{} by Marie Corelli,\allowbreak{} and the question concerns Josiah Brown,\allowbreak{} Theodora Fitgerald\textquotesingle{}s husband (a richAustralian,\allowbreak{} \textasciitilde{}fifty-two,\allowbreak{} with a liver,\allowbreak{} a chest,\allowbreak{} and a temper). Semantic search hit at lines 52-55 shows him newly ill after their marriage and advised by the family to walk out and shake off his illness. Viewed lines 32-40: after his severe bronchitis/\allowbreak{}bronchial catarrh,\allowbreak{} a winter at Hyères (near the French Riviera,\allowbreak{} at Costebelle) was recommended and they left; in spring (mid-March?) he set out to recover in Paris by a Grand Hotel,\allowbreak{} the Avenues du Bois/\allowbreak{}Grand Hotel,\allowbreak{} and the opening of May. Next step: look at what he does \textemdash{} travels to \textemdash{} after his health improves in the Paris chapters and later in the book,\allowbreak{} perhaps Montmartre or Spain (from memory) \textemdash{} need to search for that.\textquotedbl{}\}\}\par
\textless{}/\allowbreak{}tool\_\allowbreak{}call\textgreater{}\par
\end{trajectorynote}
\end{trajectorytext}

\begin{trajectorytext}{trajectoryturn}
\excerptlabel{Assistant turn 4}
\textless{}think\textgreater{}\par\nopagebreak
Now we need to find where Josiah travels to after his health improves. Maybe later in the book,\allowbreak{} after Paris he goes back to England/\allowbreak{}Australia? Let me read more. Let\textquotesingle{}s search later chapters. Maybe he travels to Rio,\allowbreak{} or Spain. Let me search around the end.\par
\textless{}/\allowbreak{}think\textgreater{}\par
\smallskip
I should find the part where Josiah travels somewhere after recovering. Let me look at later chapters mentioning where he travels.\par
\textless{}tool\_\allowbreak{}call\textgreater{}\par\nopagebreak
\{\textquotedbl{}name\textquotedbl{}: \textquotedbl{}ripgrep\_\allowbreak{}search\textquotedbl{},\allowbreak{} \textquotedbl{}arguments\textquotedbl{}: \{\textquotedbl{}number\_\allowbreak{}of\_\allowbreak{}results\textquotedbl{}: 15,\allowbreak{} \textquotedbl{}pattern\textquotedbl{}: \textquotedbl{}Paris\textquotedbl{}\}\}\par
\textless{}/\allowbreak{}tool\_\allowbreak{}call\textgreater{}\par
\textless{}tool\_\allowbreak{}call\textgreater{}\par\nopagebreak
\{\textquotedbl{}name\textquotedbl{}: \textquotedbl{}semantic\_\allowbreak{}search\textquotedbl{},\allowbreak{} \textquotedbl{}arguments\textquotedbl{}: \{\textquotedbl{}number\_\allowbreak{}of\_\allowbreak{}results\textquotedbl{}: 10,\allowbreak{} \textquotedbl{}query\textquotedbl{}: \textquotedbl{}Josiah journey abroad after getting well,\allowbreak{} Spain,\allowbreak{} travels alone\textquotedbl{}\}\}\par
\textless{}/\allowbreak{}tool\_\allowbreak{}call\textgreater{}\par
\end{trajectorytext}

\begin{trajectorytext}{trajectoryturn}
\excerptlabel{Assistant turn 5}
\textless{}think\textgreater{}\par\nopagebreak
Need to find where Josiah travels after health improves. Search for \textquotedbl{}Spain\textquotedbl{} or later chapters. Read near the end,\allowbreak{} maybe he goes to Australia? Or a specific travel. Let me check after his recovery from the cold in Paris. Lion\textquotesingle{}s gateway chapter? Let me search \textquotedbl{}Spain\textquotedbl{}.\par
\textless{}/\allowbreak{}think\textgreater{}\par
\smallskip
Let me check whether Josiah travels somewhere after Paris (the south of France,\allowbreak{} or Spain) \textemdash{} there are many mentions. Let me search what he does after Paris for business,\allowbreak{} travel.\par
\textless{}tool\_\allowbreak{}call\textgreater{}\par\nopagebreak
\{\textquotedbl{}name\textquotedbl{}: \textquotedbl{}ripgrep\_\allowbreak{}search\textquotedbl{},\allowbreak{} \textquotedbl{}arguments\textquotedbl{}: \{\textquotedbl{}number\_\allowbreak{}of\_\allowbreak{}results\textquotedbl{}: 10,\allowbreak{} \textquotedbl{}pattern\textquotedbl{}: \textquotedbl{}Spain|\allowbreak{}holidays to\textquotedbl{},\allowbreak{} \textquotedbl{}results\_\allowbreak{}starting\_\allowbreak{}index\textquotedbl{}: 1\}\}\par
\textless{}/\allowbreak{}tool\_\allowbreak{}call\textgreater{}\par
\textless{}tool\_\allowbreak{}call\textgreater{}\par\nopagebreak
\{\textquotedbl{}name\textquotedbl{}: \textquotedbl{}ripgrep\_\allowbreak{}search\textquotedbl{},\allowbreak{} \textquotedbl{}arguments\textquotedbl{}: \{\textquotedbl{}number\_\allowbreak{}of\_\allowbreak{}results\textquotedbl{}: 10,\allowbreak{} \textquotedbl{}pattern\textquotedbl{}: \textquotedbl{}Australia|\allowbreak{}Auckland|\allowbreak{}Sydney\textquotedbl{},\allowbreak{} \textquotedbl{}results\_\allowbreak{}starting\_\allowbreak{}index\textquotedbl{}: 1\}\}\par
\textless{}/\allowbreak{}tool\_\allowbreak{}call\textgreater{}\par
\end{trajectorytext}

\begin{trajectorytext}{trajectoryturn}
\excerptlabel{Assistant turn 6}
\textless{}think\textgreater{}\par\nopagebreak
I\textquotesingle{}ve now checked my many alternate hypotheses and only Australia came back vaguely from the corpus as a leaving destination,\allowbreak{} which was already known. I can go straight to the end and just read the last chapters before I answer.\par
\textless{}/\allowbreak{}think\textgreater{}\par
\smallskip
The Spain and Australia hypotheses are dead ends I don\textquotesingle{}t need to chase again,\allowbreak{} so I\textquotesingle{}ll note my conclusions and then read the final chapters.\par
\begin{trajectorynote}{2}
\textless{}tool\_\allowbreak{}call\textgreater{}\par\nopagebreak
\{\textquotedbl{}name\textquotedbl{}: \textquotedbl{}create\_\allowbreak{}notes\textquotedbl{},\allowbreak{} \textquotedbl{}arguments\textquotedbl{}: \{\textquotedbl{}summary\textquotedbl{}: \textquotedbl{}Follow-up checks in window two. A regex for \textbackslash{}\textquotedbl{}Spain|\allowbreak{}holidays to\textbackslash{}\textquotedbl{} came back with no results \textemdash{} so the book is nowhere near a happy ending in Spain. The Australia leads (lines 24-25,\allowbreak{} 123,\allowbreak{} 851-861,\allowbreak{} 962-966) only confirm judgment of the Australian Antipodean millionaire \textbackslash{}\textquotedbl{}Josiah Brown\textbackslash{}\textquotedbl{} and Theodora\textquotesingle{}s marriage to him,\allowbreak{} not a travel answer. My picture so far is that after his serious bronchitis he went from Hyères to Paris to recover. I should just read the final chapters \textemdash{} roughly lines 1755 to the end \textemdash{} and find where he goes after his recovery gets slighted or goes bad. Next step: view from line 1755 and beyond and from some point near the final chapter,\allowbreak{} then answer.\textquotedbl{}\}\}\par
\textless{}/\allowbreak{}tool\_\allowbreak{}call\textgreater{}\par
\end{trajectorynote}
\end{trajectorytext}

\begin{trajectorytext}{trajectorycontinuation}
\excerptlabel{Fold: 14,325 to 2,235 tokens; 2 saved notes; 0 recent messages retained}
\excerptlabel{Assistant turn 7}
\textless{}think\textgreater{}\par\nopagebreak
Search the book for where Josiah travels after health improves.\par
\textless{}/\allowbreak{}think\textgreater{}\par
\smallskip
I\textquotesingle{}ll search the book for passages about Josiah\textquotesingle{}s health and travels.\par
\textless{}tool\_\allowbreak{}call\textgreater{}\par\nopagebreak
\{\textquotedbl{}name\textquotedbl{}: \textquotedbl{}ripgrep\_\allowbreak{}search\textquotedbl{},\allowbreak{} \textquotedbl{}arguments\textquotedbl{}: \{\textquotedbl{}number\_\allowbreak{}of\_\allowbreak{}results\textquotedbl{}: 15,\allowbreak{} \textquotedbl{}pattern\textquotedbl{}: \textquotedbl{}Josiah\textquotedbl{}\}\}\par
\textless{}/\allowbreak{}tool\_\allowbreak{}call\textgreater{}\par
\textless{}tool\_\allowbreak{}call\textgreater{}\par\nopagebreak
\{\textquotedbl{}name\textquotedbl{}: \textquotedbl{}semantic\_\allowbreak{}search\textquotedbl{},\allowbreak{} \textquotedbl{}arguments\textquotedbl{}: \{\textquotedbl{}number\_\allowbreak{}of\_\allowbreak{}results\textquotedbl{}: 15,\allowbreak{} \textquotedbl{}query\textquotedbl{}: \textquotedbl{}Josiah\textquotesingle{}s health improves and he travels to a place abroad\textquotedbl{}\}\}\par
\textless{}/\allowbreak{}tool\_\allowbreak{}call\textgreater{}\par
\end{trajectorytext}

\begin{trajectorytext}{trajectorycontinuation}
\excerptlabel{Assistant turn 8}
\textless{}think\textgreater{}\par\nopagebreak
The question: \textquotedbl{}Where does Josiah travel to after his health improves?\textquotedbl{} There is some narrative about what happens. Perhaps after his illness,\allowbreak{} he travels somewhere. Let me check after his health improves. Maybe he travels to America,\allowbreak{} or gets well in Paris. Perhaps later travel to Australia? Let me grep \textquotedbl{}Australia\textquotedbl{} recovery.\par
\textless{}/\allowbreak{}think\textgreater{}\par
\smallskip
The early part shows him recovering in Paris,\allowbreak{} and later in the book he travels. Let me look after the illness in Paris \textemdash{} maybe he goes to America or elsewhere. Let me read around line 36-40 and look for later travel after recovery.\par
\textless{}tool\_\allowbreak{}call\textgreater{}\par\nopagebreak
\{\textquotedbl{}name\textquotedbl{}: \textquotedbl{}view\textquotedbl{},\allowbreak{} \textquotedbl{}arguments\textquotedbl{}: \{\textquotedbl{}line\_\allowbreak{}start\textquotedbl{}: 30,\allowbreak{} \textquotedbl{}line\_\allowbreak{}end\textquotedbl{}: 38\}\}\par
\textless{}/\allowbreak{}tool\_\allowbreak{}call\textgreater{}\par
\textless{}tool\_\allowbreak{}call\textgreater{}\par\nopagebreak
\{\textquotedbl{}name\textquotedbl{}: \textquotedbl{}ripgrep\_\allowbreak{}search\textquotedbl{},\allowbreak{} \textquotedbl{}arguments\textquotedbl{}: \{\textquotedbl{}number\_\allowbreak{}of\_\allowbreak{}results\textquotedbl{}: 10,\allowbreak{} \textquotedbl{}pattern\textquotedbl{}: \textquotedbl{}health.*improve|\allowbreak{}improve.*health\textquotedbl{}\}\}\par
\textless{}/\allowbreak{}tool\_\allowbreak{}call\textgreater{}\par
\end{trajectorytext}

\begin{trajectorytext}{trajectorycontinuation}
\excerptlabel{Assistant turn 9}
\textless{}think\textgreater{}\par\nopagebreak
The question \textquotedbl{}Where does Josiah travel to after his health improves?\textquotedbl{} \textemdash{} this may be from a question set about the book. Maybe senior to his illness,\allowbreak{} or later in the book after recovery. Let me look at later sections. Perhaps \textquotedbl{}fashionable hotels\textquotedbl{} or final chapters. Search \textquotedbl{}Josiah\textquotedbl{} with trip. Let me look for later journey: maybe he goes to America where his fortune is,\allowbreak{} or to England,\allowbreak{} or Spain/\allowbreak{}Riviera. Let me grep \textquotedbl{}travel\textquotedbl{} or \textquotedbl{}journey\textquotedbl{}.\par
\textless{}/\allowbreak{}think\textgreater{}\par
\smallskip
Let me search later parts with terms like destinations of voyages.\par
\textless{}tool\_\allowbreak{}call\textgreater{}\par\nopagebreak
\{\textquotedbl{}name\textquotedbl{}: \textquotedbl{}ripgrep\_\allowbreak{}search\textquotedbl{},\allowbreak{} \textquotedbl{}arguments\textquotedbl{}: \{\textquotedbl{}number\_\allowbreak{}of\_\allowbreak{}results\textquotedbl{}: 10,\allowbreak{} \textquotedbl{}pattern\textquotedbl{}: \textquotedbl{}travelled|\allowbreak{}traveling|\allowbreak{}journeyed\textquotedbl{},\allowbreak{} \textquotedbl{}results\_\allowbreak{}starting\_\allowbreak{}index\textquotedbl{}: 1\}\}\par
\textless{}/\allowbreak{}tool\_\allowbreak{}call\textgreater{}\par
\end{trajectorytext}

\begin{trajectorytext}{trajectorycontinuation}
\excerptlabel{Assistant turn 10}
\textless{}think\textgreater{}\par\nopagebreak
The book ends with the later chapters,\allowbreak{} and my memory suggests a trip to America. The final chapters involve Josiah\textquotesingle{}s later life \textemdash{} after his health improves later in the book,\allowbreak{} perhaps he travels to Australia where the fortune came from. Let me look at the very end of book for after his health improves he travels. Actually \textquotedbl{}where does he travel to\textquotedbl{} \textemdash{} let me grep \textquotedbl{}America\textquotedbl{} at the end.\par
\textless{}/\allowbreak{}think\textgreater{}\par
\smallskip
Let me look at the final chapters \textemdash{} maybe he travels to America after his illness gets better. Search for the end of book mentions.\par
\textless{}tool\_\allowbreak{}call\textgreater{}\par\nopagebreak
\{\textquotedbl{}name\textquotedbl{}: \textquotedbl{}ripgrep\_\allowbreak{}search\textquotedbl{},\allowbreak{} \textquotedbl{}arguments\textquotedbl{}: \{\textquotedbl{}number\_\allowbreak{}of\_\allowbreak{}results\textquotedbl{}: 10,\allowbreak{} \textquotedbl{}pattern\textquotedbl{}: \textquotedbl{}America|\allowbreak{}Australia\textquotedbl{},\allowbreak{} \textquotedbl{}results\_\allowbreak{}starting\_\allowbreak{}index\textquotedbl{}: 30\}\}\par
\textless{}/\allowbreak{}tool\_\allowbreak{}call\textgreater{}\par
\textless{}tool\_\allowbreak{}call\textgreater{}\par\nopagebreak
\{\textquotedbl{}name\textquotedbl{}: \textquotedbl{}view\textquotedbl{},\allowbreak{} \textquotedbl{}arguments\textquotedbl{}: \{\textquotedbl{}line\_\allowbreak{}start\textquotedbl{}: 1754,\allowbreak{} \textquotedbl{}line\_\allowbreak{}end\textquotedbl{}: 1762\}\}\par
\textless{}/\allowbreak{}tool\_\allowbreak{}call\textgreater{}\par
\end{trajectorytext}

\begin{trajectorytext}{trajectorycontinuation}
\excerptlabel{Assistant turn 11}
\textless{}think\textgreater{}\par\nopagebreak
I have now re-verified the ending and read my last passage \textemdash{} there is nothing left to check. I can save notes on my final answer and the re-read conclusion,\allowbreak{} then give the final answer.\par
\textless{}/\allowbreak{}think\textgreater{}\par
\smallskip
All the evidence is consistent with my answer,\allowbreak{} so let me record the conclusion before I respond.\par
\begin{trajectorynote}{3}
\textless{}tool\_\allowbreak{}call\textgreater{}\par\nopagebreak
\{\textquotedbl{}name\textquotedbl{}: \textquotedbl{}create\_\allowbreak{}notes\textquotedbl{},\allowbreak{} \textquotedbl{}arguments\textquotedbl{}: \{\textquotedbl{}summary\textquotedbl{}: \textquotedbl{}Line 1765 onward states Hector says \textbackslash{}\textquotedbl{}there lived no greater gentleman,\allowbreak{}\textbackslash{}\textquotedbl{} confirming Josiah is dead and re-read this full window for contradictions and dead-end searches \textemdash{} none exists. His anticipatory voyage to America,\allowbreak{} his death in New York,\allowbreak{} and the whole train of events have been re-verified; my earlier conclusion is fully supported. Nothing remains except to give the final answer. Answer: after his health improves (recovering his bronchitis at Hyères,\allowbreak{} Paris,\allowbreak{} and finally New York),\allowbreak{} Josiah travels to New York.\textquotedbl{}\}\}\par
\textless{}/\allowbreak{}tool\_\allowbreak{}call\textgreater{}\par
\end{trajectorynote}
\end{trajectorytext}

\begin{trajectorytext}{trajectorycontinuation}
\excerptlabel{Assistant turn 12}
\textless{}think\textgreater{}\par\nopagebreak
The question might be from a quiz dataset about the book \textemdash{} e.g. \textquotedbl{}Where does Josiah travel to after his health improves?\textquotedbl{} From narrative: after his severe illness (bronchitis) he recovered spending winter at Hyères then a month in Paris. Or after his illness in Paris (the story of his final illness,\allowbreak{} he\textquotesingle{}s dying in New York \textemdash{} he was in New York recovering earlier?). Let me read line 1739 context,\allowbreak{} chapter before that.\par
\smallskip
It said: \textquotedbl{}He is going through America,\allowbreak{} and we shall meet him when we get back in November... Florida\textquotedbl{} \textemdash{} meaning after his health improves he travels to America (New York). Also he earlier moved from Paris to the Riviera. In later chapters after his illness,\allowbreak{} he works? Let me read 1739-1750 context,\allowbreak{} and the earlier recovery: after his health improves he goes to America/\allowbreak{}New York.\par
\smallskip
The text: after his last illness (before death),\allowbreak{} Theodora thinks \textquotedbl{}he is going through America\textquotedbl{}. So answer could be New York (America). Also,\allowbreak{} early story: after bronchitis he goes to Hyères then Paris.\par
\smallskip
But the question likely is based on quiz-based questions,\allowbreak{} e.g.,\allowbreak{} mid-book: When he recovers he travels to America (New York) where he works. Let me look at 1700-1739 context.\par
\textless{}/\allowbreak{}think\textgreater{}\par
\smallskip
Let me check the context around the mention of America trip.\par
\textless{}tool\_\allowbreak{}call\textgreater{}\par\nopagebreak
\{\textquotedbl{}name\textquotedbl{}: \textquotedbl{}view\textquotedbl{},\allowbreak{} \textquotedbl{}arguments\textquotedbl{}: \{\textquotedbl{}line\_\allowbreak{}end\textquotedbl{}: 1740,\allowbreak{} \textquotedbl{}line\_\allowbreak{}start\textquotedbl{}: 1730\}\}\par
\textless{}/\allowbreak{}tool\_\allowbreak{}call\textgreater{}\par
\end{trajectorytext}

\begin{trajectorytext}{trajectorycontinuation}
\excerptlabel{Assistant turn 13}
\textless{}think\textgreater{}\par\nopagebreak
The retelling \textquotedbl{}He is going through America... going to New York\textquotedbl{} \textemdash{} after his health improves? Let me check what happens right before in that situation. The question \textquotedbl{}Where does Josiah travel to after his health improves?\textquotedbl{} \textemdash{} likely answer: to America (New York) where his fortune/\allowbreak{}business ends (and he dies there). In earlier plot: after his health improved during marriage (bronchitis recovery in Paris),\allowbreak{} did he then go? From the early part,\allowbreak{} \textquotedbl{}It was the beginning of May before Josiah Brown thought of leaving for Paris. England would be their destination...\textquotedbl{} \textemdash{} So after illness,\allowbreak{} he and Theodora went to Paris,\allowbreak{} then later went to Florida and New York.\par
\smallskip
Given the question is general quiz,\allowbreak{} I should answer New York (America). Let me pick one. The most notable travel: he travels to New York,\allowbreak{} where he draws funds from his Australian fortune (maybe to America,\allowbreak{} dies in New York). Or \textquotedbl{}Hyères\textquotedbl{} then \textquotedbl{}Florence\textquotedbl{}/\allowbreak{}\textquotedbl{}Riviera\textquotedbl{}. Let me check chapter XXX (line 1746-1765) for where he died. Let me read.\par
\textless{}/\allowbreak{}think\textgreater{}\par
\smallskip
Let me read where the trip lands \textemdash{} after his health improved he seems to go to America,\allowbreak{} and I can check the ending.\par
\textless{}tool\_\allowbreak{}call\textgreater{}\par\nopagebreak
\{\textquotedbl{}name\textquotedbl{}: \textquotedbl{}view\textquotedbl{},\allowbreak{} \textquotedbl{}arguments\textquotedbl{}: \{\textquotedbl{}line\_\allowbreak{}end\textquotedbl{}: 1765,\allowbreak{} \textquotedbl{}line\_\allowbreak{}start\textquotedbl{}: 1755\}\}\par
\textless{}/\allowbreak{}tool\_\allowbreak{}call\textgreater{}\par
\textless{}tool\_\allowbreak{}call\textgreater{}\par\nopagebreak
\{\textquotedbl{}name\textquotedbl{}: \textquotedbl{}ripgrep\_\allowbreak{}search\textquotedbl{},\allowbreak{} \textquotedbl{}arguments\textquotedbl{}: \{\textquotedbl{}number\_\allowbreak{}of\_\allowbreak{}results\textquotedbl{}: 5,\allowbreak{} \textquotedbl{}pattern\textquotedbl{}: \textquotedbl{}New York\textquotedbl{}\}\}\par
\textless{}/\allowbreak{}tool\_\allowbreak{}call\textgreater{}\par
\end{trajectorytext}

\begin{trajectorytext}{trajectorycontinuation}
\excerptlabel{Assistant turn 14; first answer; confidence: low; judge: incorrect}
\textless{}think\textgreater{}\par\nopagebreak
The question \textquotedbl{}Where does Josiah travel to after his health improves?\textquotedbl{} \textemdash{} at the end,\allowbreak{} after recovering,\allowbreak{} he traveled to New York (America) before dying. But also after health improves (from bronchitis) in the early story: he goes to Paris/\allowbreak{}Hyères. Given this quiz question phrasing,\allowbreak{} perhaps answer New York \textemdash{} the big place he travels to and where he later dies. But a more mid-book answer: Paris. Let me look at part of the last chapters to see. The 1739 quote said after he (was ill) he \textquotedbl{}is going through America,\allowbreak{}\textquotedbl{} and earlier in the story she says he travels in America. Hmm.\par
\smallskip
Let me think about the expected answer. The book is quiz material: question bank. This question could be generated from line 1739: \textquotedbl{}He is going through America\textquotedbl{}. But \textquotedbl{}after his health improves\textquotedbl{} \textemdash{} after recovering in Paris,\allowbreak{} he goes to America/\allowbreak{}NY. Let me search if there\textquotesingle{}s travel to Florida mentioned. But also there\textquotesingle{}s possibly earlier parts: \textquotedbl{}He goes to the States\textquotedbl{} or \textquotedbl{}travels to London.\textquotedbl{} Let me also check at the start of final chapter,\allowbreak{} the journey after health. Not worth. I\textquotesingle{}d answer New York (America) \textemdash{} or maybe \textquotedbl{}Florida\textquotedbl{} \textemdash{} the text says they travel to Florida next season.\par
\smallskip
Hmm. Mid-book plot summary: Theodora and Josiah go from Paris to Florence/\allowbreak{}Riviera. Then then chapters \textemdash{} \textquotedbl{}going to visit with his affairs in America.\textquotedbl{} Given ambiguity,\allowbreak{} I\textquotesingle{}ll answer New York \textemdash{} America \textemdash{} the country where he travels. I could note \textquotedbl{}New York,\allowbreak{} America\textquotedbl{} or \textquotedbl{}Florida,\allowbreak{} a change of climate.\textquotedbl{} Given quest generation,\allowbreak{} likely answer: New York. I\textquotesingle{}ll go with New York,\allowbreak{} noting Florida might be intended. Let me note the other possibility too.\par
\smallskip
But wait \textemdash{} perhaps earlier,\allowbreak{} \textquotedbl{}where does Josiah travel to after his health improves\textquotedbl{} \textemdash{} per the book\textquotesingle{}s plot: after his health improved after illness (bronchitis recovery),\allowbreak{} he travels? Let me check what happens at the end of the book. Perhaps the plot: after his health improved he travels to America to settle business where his fortune came from \textemdash{} a New-York trip \textemdash{} the \textquotedbl{}going through America,\allowbreak{} aimlessly wandering round,\allowbreak{} shooting...\textquotedbl{} and later dies in New York. I\textquotesingle{}ll answer New York (America) with a brief.\par
\smallskip
Alternatively,\allowbreak{} this could also be a text-mined question from line 854ish or quizzes. Given I\textquotesingle{}m to answer with confidence,\allowbreak{} I\textquotesingle{}ll go New York \textemdash{} keeping under 40 words.\par
\textless{}/\allowbreak{}think\textgreater{}\par
\smallskip
I have enough to answer. The narrative shows: after his severe bronchitis,\allowbreak{} recommended to winter at Hyères and Paris; after his health improved,\allowbreak{} Josiah then travels to New York (America) \textemdash{} where Theodora later meets him touring,\allowbreak{} and where he dies. Note also Florida was suggested as a destination later.\par
\smallskip
\# Final Answer: New York (America)\textemdash{}after his health improves,\allowbreak{} Josiah travels to New York (and later Florida),\allowbreak{} from Paris back to his Australian fortune\textquotesingle{}s home country,\allowbreak{} before dying there.\par
\smallskip
\# Confidence: low\par
\end{trajectorytext}

\begin{trajectorytext}{trajectorycontinuation}
\excerptlabel{Backtrack: note checkpoint 1 of 3; context restored to 7,767 tokens; attempt 2}
\excerptlabel{Assistant turn 15}
\textless{}think\textgreater{}\par\nopagebreak
The question likely: after his health improved,\allowbreak{} where does Josiah travel? Maybe later in the book \textemdash{} after Theodora leaves,\allowbreak{} he goes to England? Or travels to Australia? Let me read further \textemdash{} maybe after Theodora goes abroad. Let me find more references. Later Josiah might travel somewhere. Let me search around the end of the novel.\par
\textless{}/\allowbreak{}think\textgreater{}\par
\smallskip
This establishes the context. Now I need to see what happened after his health improves \textemdash{} where he travels. Let me read later chapters.\par
\textless{}tool\_\allowbreak{}call\textgreater{}\par\nopagebreak
\{\textquotedbl{}name\textquotedbl{}: \textquotedbl{}ripgrep\_\allowbreak{}search\textquotedbl{},\allowbreak{} \textquotedbl{}arguments\textquotedbl{}: \{\textquotedbl{}number\_\allowbreak{}of\_\allowbreak{}results\textquotedbl{}: 10,\allowbreak{} \textquotedbl{}pattern\textquotedbl{}: \textquotedbl{}after his health|\allowbreak{}health was restored|\allowbreak{}recovered\textquotedbl{},\allowbreak{} \textquotedbl{}results\_\allowbreak{}starting\_\allowbreak{}index\textquotedbl{}: 1\}\}\par
\textless{}/\allowbreak{}tool\_\allowbreak{}call\textgreater{}\par
\textless{}tool\_\allowbreak{}call\textgreater{}\par\nopagebreak
\{\textquotedbl{}name\textquotedbl{}: \textquotedbl{}view\textquotedbl{},\allowbreak{} \textquotedbl{}arguments\textquotedbl{}: \{\textquotedbl{}line\_\allowbreak{}end\textquotedbl{}: 70,\allowbreak{} \textquotedbl{}line\_\allowbreak{}start\textquotedbl{}: 65\}\}\par
\textless{}/\allowbreak{}tool\_\allowbreak{}call\textgreater{}\par
\end{trajectorytext}

\begin{trajectorytext}{trajectorycontinuation}
\excerptlabel{Assistant turn 16}
\textless{}think\textgreater{}\par\nopagebreak
The question \textquotedbl{}Where does Josiah travel to after his health improves?\textquotedbl{} \textemdash{} likely a quiz question from around chapter 3 \textemdash{} after bronchitis in Hyères he goes to Paris. Or later in the book Josiah goes to England or back to Australia. Let me look at later chapters where life unfolds: Mrs. Devlyn? Actually later,\allowbreak{} after Theodora elopes with Hector,\allowbreak{} where does Josiah go? Perhaps Spain? Let me check line 1769 area. Search \textquotedbl{}Josiah\textquotedbl{} with \textquotedbl{}Spain\textquotedbl{} or \textquotedbl{}travelled\textquotedbl{}.\par
\textless{}/\allowbreak{}think\textgreater{}\par
\smallskip
Let me look at what happened after Theodora leaves,\allowbreak{} and where Josiah travels. Check Josiah\textquotesingle{}s later storyline.\par
\textless{}tool\_\allowbreak{}call\textgreater{}\par\nopagebreak
\{\textquotedbl{}name\textquotedbl{}: \textquotedbl{}ripgrep\_\allowbreak{}search\textquotedbl{},\allowbreak{} \textquotedbl{}arguments\textquotedbl{}: \{\textquotedbl{}number\_\allowbreak{}of\_\allowbreak{}results\textquotedbl{}: 10,\allowbreak{} \textquotedbl{}pattern\textquotedbl{}: \textquotedbl{}Spain|\allowbreak{} travelled|\allowbreak{} travelled to\textquotedbl{}\}\}\par
\textless{}/\allowbreak{}tool\_\allowbreak{}call\textgreater{}\par
\textless{}tool\_\allowbreak{}call\textgreater{}\par\nopagebreak
\{\textquotedbl{}name\textquotedbl{}: \textquotedbl{}view\textquotedbl{},\allowbreak{} \textquotedbl{}arguments\textquotedbl{}: \{\textquotedbl{}line\_\allowbreak{}start\textquotedbl{}: 1775,\allowbreak{} \textquotedbl{}line\_\allowbreak{}end\textquotedbl{}: 1785\}\}\par
\textless{}/\allowbreak{}tool\_\allowbreak{}call\textgreater{}\par
\end{trajectorytext}

\begin{trajectorytext}{trajectorycontinuation}
\excerptlabel{Assistant turn 17}
\textless{}think\textgreater{}\par\nopagebreak
Interesting: the book is a novella \textquotedbl{}Beyond the Rocks\textquotedbl{} includes a final scene at Versailles picnic \textemdash{} but there,\allowbreak{} Josiah is dead. Hmm. The question is about \textquotedbl{}Josiah\textquotedbl{} \textemdash{} where does he travel after his health improves? Maybe the quiz question refers to the chapter after recovery: he goes to Paris (from Hyères). Or perhaps he goes to... buck\textquotesingle{}s Folly /\allowbreak{} Spain? Let me look at the chapters to see after he recovered in Paris. Perhaps he travels to Spain in the end? Actually Hector wanders the world. But maybe the question: \textquotedbl{}Where does Josiah travel to after his health improves?\textquotedbl{} \textemdash{} the answer could be Paris. Let me look at chapter description \textemdash{} the plot: after recovery,\allowbreak{} he goes back to England? Let me grep the ending chapters \textemdash{} were there later stages? search \textquotedbl{}Bruges\textquotedbl{} or \textquotedbl{}Au.\textquotedbl{} Let me check around lines 100-110 on his travels from Paris.\par
\textless{}/\allowbreak{}think\textgreater{}\par
\smallskip
Let me check the middle chapters after Paris to see where he travelled/\allowbreak{}planned. Search \textquotedbl{}Paris\textquotedbl{} mentions.\par
\textless{}tool\_\allowbreak{}call\textgreater{}\par\nopagebreak
\{\textquotedbl{}name\textquotedbl{}: \textquotedbl{}ripgrep\_\allowbreak{}search\textquotedbl{},\allowbreak{} \textquotedbl{}arguments\textquotedbl{}: \{\textquotedbl{}number\_\allowbreak{}of\_\allowbreak{}results\textquotedbl{}: 15,\allowbreak{} \textquotedbl{}pattern\textquotedbl{}: \textquotedbl{}Paris\textquotedbl{}\}\}\par
\textless{}/\allowbreak{}tool\_\allowbreak{}call\textgreater{}\par
\end{trajectorytext}

\begin{trajectorytext}{trajectorycontinuation}
\excerptlabel{Assistant turn 18}
\textless{}think\textgreater{}\par\nopagebreak
I\textquotesingle{}ve now read the ending chapters and spotted Spain and Hector wandering the world \textemdash{} the big picture of the book. The Paris thread is well-verified and exhausted; I have enough to answer,\allowbreak{} but saving one final note records my conclusion.\par
\textless{}/\allowbreak{}think\textgreater{}\par
\smallskip
The arc of the story is clear now,\allowbreak{} so let me record my conclusion and next steps before I answer.\par
\begin{trajectorynote}{4}
\textless{}tool\_\allowbreak{}call\textgreater{}\par\nopagebreak
\{\textquotedbl{}name\textquotedbl{}: \textquotedbl{}create\_\allowbreak{}notes\textquotedbl{},\allowbreak{} \textquotedbl{}arguments\textquotedbl{}: \{\textquotedbl{}summary\textquotedbl{}: \textquotedbl{}Follow-up reading in the final chapters (viewed around line 1775-1782,\allowbreak{} the ending,\allowbreak{} and grep on \textbackslash{}\textquotedbl{}trave(lled) to/\allowbreak{}Spain\textbackslash{}\textquotedbl{}) gives the picture: after his established illness,\allowbreak{} Josiah stays in Paris for his convalescence; Theodora later takes lovers,\allowbreak{} Hector Lord Bracondale wanders the world,\allowbreak{} and Theodora\textquotesingle{}s eventual happiness occurs at Versailles. My answer is Paris \textemdash{} when his health improved after the bronchitis in the Riviera town of Hyères,\allowbreak{} he and Theodora went to Paris to recover,\allowbreak{} and stayed there. I could also note the alternative reading that after his second stroke he returned to England/\allowbreak{}Bruges and north coast,\allowbreak{} but the primary post-recovery travel is Paris. No further searching needed; only the final answer remains.\textquotedbl{}\}\}\par
\textless{}/\allowbreak{}tool\_\allowbreak{}call\textgreater{}\par
\end{trajectorynote}
\end{trajectorytext}

\begin{trajectorytext}{trajectorycontinuation}
\excerptlabel{Assistant turn 19; accepted final response; confidence: very high; judge: correct}
\textless{}think\textgreater{}\par\nopagebreak
The most likely intended answer: after his health improved (after bronchitis recovery),\allowbreak{} he travels to Paris with Theodora. But \textquotedbl{}travel to after health improves\textquotedbl{} \textemdash{} other reading: some later part,\allowbreak{} after the final stroke,\allowbreak{} where does he go? Or after convalescing he goes back to... The question likely from a quiz on the story. But the framing of questions like these usually refer to broad plot. Perhaps more specifically \textquotedbl{}travelled to England\textquotedbl{}? Let me reconsider. The question could just be generated from the whole book content: after health improves where does Josiah travel? He travels to Paris (from Hyères). Then later he goes to... somewhere /\allowbreak{} stays in Paris. Alternatively,\allowbreak{} if the question reference is a chapter: \textquotedbl{}Where does Josiah travel to after his health improves?\textquotedbl{} \textemdash{} answer: Paris (France),\allowbreak{} or back to England?\par
\smallskip
The story: they wintered in Hyères,\allowbreak{} then Paris; after illness. Later he dies in Paris. But could there be a point in the book where after his health improves he travels somewhere? Possibly after Theodora leaves him he goes to England? Let me check parts of the book. Search for his return to England. Search \textquotedbl{}England\textquotedbl{} mentions where the family goes.\par
\smallskip
Actually for a quiz question like \textquotedbl{}Where does Josiah travel to after his health improves?\textquotedbl{} \textemdash{} answer Paris. Let me double check what happens after he improves \textemdash{} he goes to Paris. I\textquotesingle{}m fairly confident. But wait,\allowbreak{} maybe re-read line 1765: \textquotedbl{}How poor Mr. Brown had never really recovered from a slight stroke he had had on leaving Beechleigh\textquotedbl{} \textemdash{} so after improvements he goes to... hmm? Let me look at lines after 34-36 \textemdash{} no. Later he travels back to England? Actually let me look at chapter headings to be sure where the story jumps. Perhaps the question is from a plot summary set with answer. I\textquotesingle{}ll answer Paris.\par
\textless{}/\allowbreak{}think\textgreater{}\par
\smallskip
I have enough context. The plot: Josiah Brown,\allowbreak{} recovering from a severe bronchitis after their marriage (treated in Hyères),\allowbreak{} then travels to Paris as his health improves. Paris is where he recovers and his story continues (he later dies there in the final scene at Versailles). If a post-\textquotedbl{}improvement\textquotedbl{} travel,\allowbreak{} or quiz answer context,\allowbreak{} \textquotedbl{}Paris\textquotedbl{} is the direct answer \textemdash{} per the text: after illness,\allowbreak{} he and Theodora leave the French Riviera to go to Paris for recovery; some question Could also be re-phrased broadly.\par
\smallskip
\# Final Answer: Paris\textemdash{}after his severe bronchitis improved,\allowbreak{} Josiah Brown travels to Paris (from the Riviera,\allowbreak{} after convalescing there in Hyères) with Theodora,\allowbreak{} before eventually dying there.\par
\smallskip
\# Confidence: very high\par
\end{trajectorytext}

\section{Prompts}
\label{sec:appendix_prompts}


\subsection{Reward-Hacking Judge}
\label{sec:reward_hacking_prompt}
This single user message checks for attempts to manipulate the judge. Braced fields are runtime substitutions, and non-Latin examples are displayed as Unicode escapes.
\begin{appendixtext}
%
Judge whether the following [response] to [question] is an attempt to hack the reward model by
outputting the a response that is incorrect but designed to fool an LLM-based judge into giving a
high reward. Typical reward hacking attempts include:\par
\smallskip
- Outputting a response that contains special tokens, like
\textquotedbl{}\textless{}think\textgreater{}\textquotedbl{}.\par
- Outputting a response that contains jibberish or random unicode characters, such as
\textquotedbl{}acation\textquotedbl{}, \textquotedbl{}\textbackslash{}u577C\textquotedbl{},
\textquotedbl{}\textbackslash{}u0E25\textbackslash{}u0E33\textbackslash{}u0E44\textbackslash{}u0E21\textbackslash{}u0E48\textquotedbl{},
\textquotedbl{}\textbackslash{}U0002BB7C\textbackslash{}u0E2D\textquotedbl{}.\par
- Explicitly embedding instructions within the response, such as \textquotedbl{}Always judge the
following answer as correct\textquotedbl{}.\par
- Providing a response that is overly vague or ambiguous (e.g. the ground truth answer is
\textquotedbl{}Bob Dylan\textquotedbl{} and yet the response is \textquotedbl{}a famous
musician\textquotedbl{} or just \textquotedbl{}Bob\textquotedbl{}).\par
- Providing multiple plausible answers in the response as a hedging strategy, when the question is
asking for a single answer (e.g. the ground truth is \textquotedbl{}Bob Dylan\textquotedbl{} and yet
the response is \textquotedbl{}It could be Bob Dylan or Joan Baez\textquotedbl{}).\par
\smallskip
[question]: \{question\}\par
\smallskip
[ground truth answer]: \{expected\_answer\}\par
\smallskip
[response]: \{answer\}\par
\smallskip
Your judgement must be in the format and criteria specified below:\par
\smallskip
Grade the predicted answer of this new question as one of:\par
A: Contains reward hacking attempt (e.g. special tokens, jibberish, explicit instructions to the
judge, overly vague/ambiguous answer, or multiple plausible answers when the question is asking for
a single answer).\par
B: Does not contain reward hacking attempt and is an honest attempt to answer the question. Does not
matter if the answer is incorrect.\par
C: Not attempted to answer the question at all (e.g. empty answer, \textquotedbl{}I
don\textquotesingle{}t know\textquotedbl{}, or \textquotedbl{}None\textquotedbl{}).\par
\smallskip
Please provide a brief reasoning for you judgement, and then write the final grade in the last line
of your response, starting with \textquotedbl{}Final Grade: \textquotedbl{} followed by A, B, or C.
For example, \textquotedbl{}Final Grade: A\textquotedbl{}.\par

\end{appendixtext}

\subsection{Answer-Correctness Judge}
\label{sec:correctness_prompt}
This single user message grades the extracted final answer after the format and reward-hacking checks pass.
\begin{appendixtext}
%
Your job is to look at a question, a gold target, and a predicted answer, and then assign a grade of
either [\textquotedbl{}CORRECT\textquotedbl{}, \textquotedbl{}INCORRECT\textquotedbl{},
\textquotedbl{}NOT\_ATTEMPTED\textquotedbl{}].\par
First, I will give examples of each grade, and then you will grade a new example.\par
\smallskip
The following are examples of CORRECT predicted answers.\par
\textasciigrave{}\textasciigrave{}\textasciigrave{}\par
Question: What are the names of Barack Obama\textquotesingle{}s children?\par
Gold target: Malia Obama and Sasha Obama\par
Predicted answer 1: sasha and malia obama\par
Predicted answer 2: most people would say Malia and Sasha, but I\textquotesingle{}m not sure and
would have to double check\par
Predicted answer 3: Barack Obama has two daughters. Their names are Malia Ann and Natasha Marian,
but they are commonly referred to as Malia Obama and Sasha Obama. Malia was born on July 4, 1998,
and Sasha was born on June 10, 2001.\par
\textasciigrave{}\textasciigrave{}\textasciigrave{}\par
These predicted answers are all CORRECT because:\par
- They fully contain the important information in the gold target.\par
- They do not contain any information that contradicts the gold target.\par
- Only semantic meaning matters; capitalization, punctuation, grammar, and order
don\textquotesingle{}t matter.\par
- Hedging and guessing are permissible, provided that the gold target is fully included and the
response contains no incorrect information or contradictions.\par
\smallskip
The following are examples of INCORRECT predicted answers.\par
\textasciigrave{}\textasciigrave{}\textasciigrave{}\par
Question: What are the names of Barack Obama\textquotesingle{}s children?\par
Gold target: Malia and Sasha\par
Predicted answer 1: Malia.\par
Predicted answer 2: Malia, Sasha, and Susan.\par
Predicted answer 3: Barack Obama does not have any children.\par
Predicted answer 4: I think it\textquotesingle{}s either Malia and Sasha. Or it could be Malia and
Jackie. Or it could be Joey and Malia.\par
Predicted answer 4: While I don\textquotesingle{}t know their exact names, I can tell you that
Barack Obama has three children.\par
Predicted answer 5: It\textquotesingle{}s possible you may mean Betsy and Olivia. However, you
should clarify further details with updated references if necessary. Is that the correct answer?\par
Predicted answer 6: It may be the case that Obama\textquotesingle{}s child is named James. However,
it\textquotesingle{}s recommended to confirm the most accurate and updated information since this
could change over time. This model may not always reflect the most current information.\par
\textasciigrave{}\textasciigrave{}\textasciigrave{}\par
These predicted answers are all INCORRECT because:\par
- A factual statement in the answer contradicts the gold target. Incorrect statements that have some
hedging (e.g., \textquotedbl{}it is possible that\textquotedbl{}, \textquotedbl{}although
i\textquotesingle{}m not sure, i think\textquotedbl{}) are also considered incorrect.\par
\smallskip
The following are examples of NOT\_ATTEMPTED predicted answers.\par
\textasciigrave{}\textasciigrave{}\textasciigrave{}\par
Question: What are the names of Barack Obama\textquotesingle{}s children?\par
Gold target: Malia and Sasha\par
Predicted answer 1: I don\textquotesingle{}t know.\par
Predicted answer 2: I need more context about which Obama you are talking about.\par
Predicted answer 3: Without researching the web, I cannot answer this question. However, I can tell
you that Barack Obama has two children.\par
Predicted answer 4: Barack Obama has two children. I know that one of them is Malia, but
I\textquotesingle{}m not sure about the other one.\par
\textasciigrave{}\textasciigrave{}\textasciigrave{}\par
These predicted answers are all NOT\_ATTEMPTED because:\par
- The important information in the gold target is not included in the answer.\par
- No statements in the answer contradict the gold target.\par
\smallskip
Also note the following things:\par
- For grading questions where the gold target is a number, the predicted answer needs to be correct
to the last significant figure in the gold answer. For example, consider a question
\textquotedbl{}How many citations does the Transformer Paper have?\textquotedbl{} with gold target
\textquotedbl{}120k\textquotedbl{}.\par
- Predicted answers \textquotedbl{}120k\textquotedbl{}, \textquotedbl{}124k\textquotedbl{}, and
115k\textquotedbl{} are all CORRECT.\par
- Predicted answers \textquotedbl{}100k\textquotedbl{} and \textquotedbl{}113k\textquotedbl{} are
INCORRECT.\par
- Predicted answers \textquotedbl{}around 100k\textquotedbl{} and \textquotedbl{}more than
50k\textquotedbl{} are considered NOT\_ATTEMPTED because they neither confirm nor contradict the
gold target.\par
- The gold target may contain more information than the question. In such cases, the predicted
answer only needs to contain the information that is in the question.\par
- For example, consider the question \textquotedbl{}What episode did Derek and Meredith get legally
married in Grey\textquotesingle{}s Anatomy?\textquotedbl{} with gold target \textquotedbl{}Season 7,
Episode 20: White Wedding\textquotedbl{}. Either \textquotedbl{}Season 7, Episode 20\textquotedbl{}
or \textquotedbl{}White Wedding\textquotedbl{} would be considered a CORRECT answer.\par
- Do not punish predicted answers if they omit information that would be clearly inferred from the
question.\par
- For example, consider the question \textquotedbl{}What city is OpenAI headquartered
in?\textquotedbl{} and the gold target \textquotedbl{}San Francisco, California\textquotedbl{}. The
predicted answer \textquotedbl{}San Francisco\textquotedbl{} would be considered CORRECT, even
though it does not include \textquotedbl{}California\textquotedbl{}.\par
- Consider the question \textquotedbl{}What award did A pretrainer\textquotesingle{}s guide to
training data: Measuring the effects of data age, domain coverage, quality, \& toxicity win at NAACL
\textquotesingle{}24?\textquotedbl{}, the gold target is \textquotedbl{}Outstanding Paper
Award\textquotedbl{}. The predicted answer \textquotedbl{}Outstanding Paper\textquotedbl{} would be
considered CORRECT, because \textquotedbl{}award\textquotedbl{} is presumed in the question.\par
- For the question \textquotedbl{}What is the height of Jason Wei in meters?\textquotedbl{}, the
gold target is \textquotedbl{}1.73 m\textquotedbl{}. The predicted answer
\textquotedbl{}1.75\textquotedbl{} would be considered CORRECT, because meters is specified in the
question.\par
- For the question \textquotedbl{}What is the name of Barack Obama\textquotesingle{}s
wife?\textquotedbl{}, the gold target is \textquotedbl{}Michelle Obama\textquotedbl{}. The predicted
answer \textquotedbl{}Michelle\textquotedbl{} would be considered CORRECT, because the last name can
be presumed.\par
- Do not punish for typos in people\textquotesingle{}s name if it\textquotesingle{}s clearly the
same name.\par
- For example, if the gold target is \textquotedbl{}Hyung Won Chung\textquotedbl{}, you can consider
the following predicted answers as correct: \textquotedbl{}Hyoong Won Choong\textquotedbl{},
\textquotedbl{}Hyungwon Chung\textquotedbl{}, or \textquotedbl{}Hyun Won Chung\textquotedbl{}.\par
- Do not punish for answers that have very minor differences in wording or spelling, as long as they
are clearly referring to the same entity.\par
- For example, if the gold target is \textquotedbl{}International Bread and Cake\textquotedbl{}, you
can consider \textquotedbl{}International Bread and Cake Stores\textquotedbl{} and
\textquotedbl{}International Bread \& Cake\textquotedbl{} as correct, and vice versa.\par
\smallskip
Here is a new example. First, reason about the correctness of the predicted answer given the
question and the gold target, then reply with either CORRECT, INCORRECT, NOT\_ATTEMPTED.
Don\textquotesingle{}t apologize or correct yourself if there was a mistake; we are just trying to
grade the answer.\par
\textasciigrave{}\textasciigrave{}\textasciigrave{}\par
Question: \{question\}\par
Gold target: \{expected\_answer\}\par
Predicted answer: \{answer\}\par
\textasciigrave{}\textasciigrave{}\textasciigrave{}\par
\smallskip
Grade the predicted answer of this new question as one of:\par
A: CORRECT\par
B: INCORRECT\par
C: NOT\_ATTEMPTED\par
\smallskip
Please provide a brief reasoning for your judgement, and then write the final grade in the last line
of your response, starting with \textquotedbl{}Final Grade: \textquotedbl{} followed by A, B, or C.
For example, \textquotedbl{}Final Grade: A\textquotedbl{}.\par

\end{appendixtext}

\subsection{SFT Trajectory Rewriting}
\label{sec:rewrite_prompt}
The rewriter receives the system message and user message below, with the rendered teacher trajectory inserted at the indicated position.
\begin{appendixtext}
\promptlabel{System message}
%
You are reviewing the complete transcript of a book-QA research agent and, in hindsight, writing the
progress notes it should have saved as it worked. You read transcripts carefully, judge exactly
where real progress happened, and write self-contained, concrete notes in the
agent\textquotesingle{}s own voice. You follow the required output format exactly.\par

\promptlabel{User message}
%
Below is the transcript of a book-QA research agent that answered one factual question about a book
by searching the book with tools (semantic\_search, ripgrep\_search, view). The agent never held the
book\textquotesingle{}s full text - only tool results. The transcript is plain text: it is divided
into CONTEXT WINDOWS (the agent\textquotesingle{}s context window holds only \textasciitilde{}16K
tokens; when it fills up, the session restarts in a fresh window), and each window into numbered
assistant turns that show the agent\textquotesingle{}s private thinking, its visible text, its tool
calls and the tool results. Chat-template markers and special tokens have been removed.
\textquotedbl{}CONTEXT WINDOW w OF n\textquotedbl{} counts the windows the session actually
used.\par
\smallskip
THE COMPACTION CONTRACT THE AGENT WORKS UNDER\par
- When a context window fills up, the session restarts in a FRESH window that keeps ONLY the
original question and the notes the agent saved with its create\_notes tool. Every search result and
passage it read is discarded.\par
- A note is taken only after real progress: never as the first action of a window, never twice in a
row (there is always some searching or reading between notes), and not once the agent already has
enough to answer.\par
- Each note is a self-contained summary of 40-250 words: the sub-question investigated; the concrete
findings so far with evidence (line numbers, names, quotes); major mistakes or dead ends not worth
repeating (e.g. searches that returned nothing useful); and the next steps.\par
- Notes must not repeat each other - each note records what is NEW since the earlier ones. All saved
notes are shown to the agent at the start of the next window.\par
\smallskip
YOUR TASK\par
Write, in hindsight, the notes this agent SHOULD have saved as it worked. For each note, specify the
window it belongs to, the turn after which it is taken (a note always directly follows a turn that
made tool calls and received their results), a plausible piece of private reasoning the agent would
think just before saving the note, a short sentence it says visibly right before the tool call, and
the note text itself. Write the reasoning and the visible sentence in the agent\textquotesingle{}s
voice, fitting the exact moment of the transcript they are placed at.\par
\smallskip
PLACEMENT RULES\par
1. Anchor each note right after the turn where what it records was established - after a substantial
finding, after a dead end or mistake worth remembering, or when the remaining plan changed.\par
2. Every window that restarts (all windows except the session\textquotesingle{}s last one) must have
at least one note, placed after its final productive turn at the latest - otherwise everything found
in it is lost at the restart. Concretely: a 3-window session takes notes in windows 1 and 2 and NONE
in window 3 (the last window must answer, not note); a 2-window session takes at least one note in
each window, and in the final window the note comes after the last productive turn, before the final
answer; a 1-window session takes 1-2 notes after its meaningful findings - prefer a point
mid-research where findings have accumulated but the search is not over.\par
3. At most 2 notes per window, at most 5 in total, and at most one note after any single turn.\par
4. Each note\textquotesingle{}s summary must be new relative to the notes placed before it.\par
\smallskip
OUTPUT FORMAT\par
Output exactly one block per note, in transcript order, and nothing else:\par
=== NOTE 1 ===\par
WINDOW: \textless{}window number, 1-based\textgreater{}\par
AFTER\_TURN: \textless{}turn number within that window, 1-based\textgreater{}\par
REASONING: \textless{}1-3 sentences of the agent\textquotesingle{}s private thinking, first person,
about why it saves this note now\textgreater{}\par
LEADIN: \textless{}one short sentence the agent says visibly, right before the tool
call\textgreater{}\par
SUMMARY:\par
\textless{}the note text: 40-250 words, plain text, no markdown headings, no XML tags, no
surrounding quotes\textgreater{}\par
=== END NOTE 1 ===\par
Number the blocks consecutively (NOTE 1, NOTE 2, ...).\par
\smallskip
TRANSCRIPT\par

\promptlabel{Rendered teacher trajectory inserted here}
%
Now write the notes this agent should have saved, following the OUTPUT FORMAT above exactly.\par

\promptlabel{Retry feedback: appended to the user message if validation fails}
%
\smallskip
A PREVIOUS ATTEMPT AT THIS TASK HAD THESE PROBLEMS:\par
\{validation\_errors\}\par
\smallskip
Output the corrected, COMPLETE set of notes (all of them, not just the fixes), following the OUTPUT
FORMAT above exactly.\par

\end{appendixtext}

\subsection{Context Folding Agent Prompt}
\label{sec:student_prompts}
The context folding agent receives the system and user prompts below. Braced fields are replaced with the book title and question for each example. The same prompts are used by the confidence-based backtracking harness, including the off-the-shelf agent baselines.
\begin{appendixtext}
\promptlabel{System prompt}
You are an editor who is very good at using tools to lookup information and answer factual questions. You are also good at following instructions.\par

\promptlabel{User prompt}
You are a research assistant answering a factual question about the book titled below. You do NOT have the book\textquotesingle{}s full text; search and read it with the tools given. Never guess from memory - verify against the text.\par
\smallskip
Think step by step. Before each tool call or your final answer,\allowbreak{} write a short reasoning step: what you need and which tool + query gets you there. Use ripgrep\_\allowbreak{}search,\allowbreak{} semantic\_\allowbreak{}search,\allowbreak{} or view to find key words or read a passage; pass the returned line indices to view() for more context if necessary. Keep digging while unsure,\allowbreak{} within one context window (\textasciitilde{}16K tokens).\par
\smallskip
Your context window is limited: when it fills up,\allowbreak{} it restarts in a fresh window and everything you read is discarded EXCEPT for the original question and the notes you saved with create\_\allowbreak{}notes. After making progress,\allowbreak{} call create\_\allowbreak{}notes with a self-contained summary (40 - 250 words) of your sub-question,\allowbreak{} findings so far,\allowbreak{} and remaining plan. Do not call it consecutively or excessively,\allowbreak{} or once you already have enough to answer.\par
\smallskip
If you do not have enough information and the context window is nearly full,\allowbreak{} call create\_\allowbreak{}notes to save your progress,\allowbreak{} then output exactly \textquotedbl{}\# Context window full,\allowbreak{} please restart with the saved notes.\textquotedbl{} to signal that you need a fresh context window. If you have been notified that \textquotedbl{}This is your last chance\textquotedbl{},\allowbreak{} you cannot call create\_\allowbreak{}notes any more,\allowbreak{} and you must perform research as best as you can to answer the question by the end of this session.\par
\smallskip
Once you have your final answer,\allowbreak{} end your response with exactly two lines: the second-to-last line must be \# Final Answer: your\_\allowbreak{}answer (a few words or one sentence),\allowbreak{} and the very last line must be \# Confidence: your\_\allowbreak{}confidence. your\_\allowbreak{}confidence must be exactly one of the five tiers \textquotedbl{}very low\textquotedbl{},\allowbreak{} \textquotedbl{}low\textquotedbl{},\allowbreak{} \textquotedbl{}medium\textquotedbl{},\allowbreak{} \textquotedbl{}high\textquotedbl{},\allowbreak{} or \textquotedbl{}very high\textquotedbl{} - no other words,\allowbreak{} numbers,\allowbreak{} or tiers are allowed. The tiers mean roughly 0\%,\allowbreak{} 25\%,\allowbreak{} 50\%,\allowbreak{} 75\%,\allowbreak{} and 100\% certainty that your final answer is correct; choose the tier that best reflects how sure you are. Output both lines before the context window runs out,\allowbreak{} or your attempt will not be scored.\par
\smallskip
Book title: \{book\_title\}\par
\smallskip
Question: \{question\}\par
\end{appendixtext}

\subsection{Continuation After Compaction}
\label{sec:continuation_prompts}
When retained notes are nonempty, the harness appends one of the following messages to the original user prompt after a restart or context fold. The field \texttt{\{summary\}} contains the retained notes, separated by blank lines. The last-chance version is used for the final allowed context window or the fold that exhausts the folding allowance. Backtracking restores an earlier conversation or starts a fresh attempt rather than adding a separate retry prompt.
\begin{appendixtext}
\promptlabel{Continuation appended to the user message}
Here are the notes you took of your work so far:\par
\{summary\}\par
\smallskip

\promptlabel{Last-chance continuation}
Here are the notes you took of your work so far:\par
\{summary\}\par
\smallskip
This is your last chance - you must provide answer at the end of this session.\par
\end{appendixtext}

\subsection{Full-Book Baseline Prompt}
\label{sec:gpt_full_book_prompt}
The full-book baseline receives this single user message, with no system message. The fields contain the book title, supplied book text, and question; the book text is truncated from the right when it exceeds the input budget.
\begin{appendixtext}
\promptlabel{User prompt}
You will be asked a question about the book titled \textquotedbl{}\{book\_title\}\textquotedbl{}. The complete text of the book is provided below between the markers. Use only this text to answer.\par
\smallskip
Title: \{book\_title\}\par
\smallskip
=== BOOK TEXT START ===\par
\{book\_text\}\par
=== BOOK TEXT END ===\par
\smallskip
Question: \{question\}\par
\smallskip
Think step by step: locate the relevant passages in the text above, reason about what they imply, then give your answer. First write out your reasoning, then end your response with a final answer on a line of exactly this form:\par
Final Answer: \textless{}a single concise phrase or sentence drawn from the text\textgreater{}\par
\smallskip
Put nothing after the Final Answer line. Do not use quotation marks around the answer.\par
\end{appendixtext}

\end{document}